\documentclass{article}

\PassOptionsToPackage{numbers, compress}{natbib}
 \usepackage[preprint]{neurips_2026}

\usepackage[utf8]{inputenc} 
\usepackage[T1]{fontenc}    
\usepackage{hyperref}       
\usepackage{url}            
\usepackage{booktabs}       
\usepackage{amsfonts}       
\usepackage{nicefrac}       
\usepackage{microtype}      

\usepackage[table]{xcolor}
\usepackage{graphicx}
\usepackage{wrapfig}
\usepackage{subcaption}

\usepackage{makecell}
\usepackage{booktabs}
\usepackage{arydshln}
\usepackage{multirow}
\usepackage{natbib}
\usepackage{amsmath}
\usepackage{algorithm}
\usepackage{algpseudocode}
\usepackage{amssymb}
\usepackage{pifont}
\usepackage[most]{tcolorbox}
\usepackage{capt-of}
\usepackage{enumitem}
\usepackage{tabularx}
\usepackage{array}
\usepackage[capitalize,noabbrev,nameinlink]{cleveref}

\newcommand{\cmark}{\color{teal}{\ding{51}}}
\newcommand{\xmark}{\color{red}{\ding{55}}}

\definecolor{myblue}{HTML}{E8F0FF} 
\definecolor{mypink}{HTML}{FDECEC}
\definecolor{mygray}{HTML}{F2F2F2}
\definecolor{myyellow}{HTML}{FFF9E6}
\definecolor{wgpink}{HTML}{F8CFCF}
\definecolor{wgblue}{HTML}{CFE0FF}
\definecolor{wbgray}{HTML}{E6E6E6}
\definecolor{mygreen}{HTML}{2ECC71}
\definecolor{myred}{HTML}{FF4D3A}

\newtcblisting{promptbox}[1][]{%
  enhanced,
  breakable,
  colback=mygray,
  colframe=black!65,
  boxrule=0.4pt,
  arc=2pt,
  left=6pt,
  right=6pt,
  top=6pt,
  bottom=6pt,
  listing only,
  listing options={
    basicstyle=\ttfamily\small,
    breaklines=true,
    columns=fullflexible
  },
  #1
}
\newtcblisting{promptboxfootnotesize}[1][]{%
  enhanced,
  breakable,
  colback=mygray,
  colframe=black!65,
  boxrule=0.4pt,
  arc=2pt,
  left=6pt,
  right=6pt,
  top=6pt,
  bottom=6pt,
  listing only,
  listing options={
    basicstyle=\ttfamily\footnotesize,
    breaklines=true,
    columns=fullflexible
  },
  #1
}
\newtcblisting{promptboxscriptsize}[1][]{%
  enhanced,
  breakable,
  colback=mygray,
  colframe=black!65,
  boxrule=0.4pt,
  arc=2pt,
  left=6pt,
  right=6pt,
  top=6pt,
  bottom=6pt,
  listing only,
  listing options={
    basicstyle=\ttfamily\scriptsize,
    breaklines=true,
    columns=fullflexible
  },
  #1
}
\definecolor{citeblue}{rgb}{0.15,0.35,0.65}
\hypersetup{
    colorlinks = true,
    linkcolor = citeblue,
    citecolor = citeblue,
    urlcolor = citeblue
}

\title{
Learn from Weaknesses: Automated Domain Specialization for Small Computer-Use Agents
}

\author{%
  Suji Kim$^{1,2}$\thanks{Equal Contribution \hfill \textbf{Project page:} {\url{https://learnweak.github.io/}}} \qquad Kangsan Kim$^{1}$\footnotemark[1] \qquad Sung Ju Hwang$^{1,3}$ \\[0.4em]
  $^1$KAIST \qquad $^2$Samsung Electronics \qquad $^3$DeepAuto.ai \\[0.2em]
  \texttt{\{suji.kim, kangsan.kim, sungju.hwang\}@kaist.ac.kr} 
}

\begin{document}

\maketitle

\newcommand{\modelname}[1]{\textsc{LearnWeak}}

\begin{abstract}
\label{sec:abstract}
Computer-use agents (CUAs) have recently made substantial progress, but deploying a separate large expert for each software domain remains expensive. 
Small open CUAs are more practical specialization targets, but they remain substantially weaker and exhibit uneven domain-specific failures. 
A straightforward remedy is to synthesize large-scale training data for the target domain, yet we find that this naive approach yields only marginal improvements.
Building on this observation, we introduce \modelname{}, an annotation-free specialization framework for small CUAs that uses a stronger reference agent to identify the student's weaknesses in the target domain, synthesize targeted tasks, and construct supervision automatically. 
\modelname{} further introduces an error-aware specialization objective that disentangles planning and execution errors, enabling more behaviorally precise updates than broad uniform supervision. 
On OSWorld, \modelname{} achieves average gains of 11.6 and 11.1 percentage points over EvoCUA-8B and OpenCUA-7B, respectively, across eight domains.
We also validate that our student-aware dataset generation and training approaches outperform existing autonomous trajectory generation and training baselines.  
Our work highlights the importance of student-awareness in both data synthesis and agent training, pointing toward a more principled and efficient path for specializing small CUAs in diverse domains.
\end{abstract}

\section{Introduction}
\label{sec:introduction}
Computer-use agents (CUAs) have advanced rapidly across desktop and web environments, with two dominant paradigms emerging: large proprietary models such as Claude Sonnet 4.6~\cite{anthropic2026sonnet46} and GPT-5.4~\cite{gpt5_4}, and small models fine-tuned specifically for computer-use tasks, such as EvoCUA~\cite{evocua} and OpenCUA~\cite{opencua}. 
The latter paradigm~\cite{os-atlas, yang2025ferret, uitarsone} is particularly compelling for real-world deployment, as fine-tuned small models enable faster and more cost-efficient inference while remaining viable for edge devices~\cite{xu2026mobileagentv35, showui} and privacy-sensitive enterprises where proprietary APIs are prohibited~\cite{erdogan-etal-2024-tinyagent, zharmagambetov2025agentdam}.
However, a substantial performance gap persists between closed models and small CUAs, particularly in domain-specific software environments with unique conventions or unfamiliar workflows~\cite{screenshot-pro, yang2025macosworld, xie2024osworld}.
Addressing this gap is therefore critical for advancing the practical deployment of small CUAs.

\begin{figure}[t]
    \centering
    \includegraphics[width=0.95\linewidth]{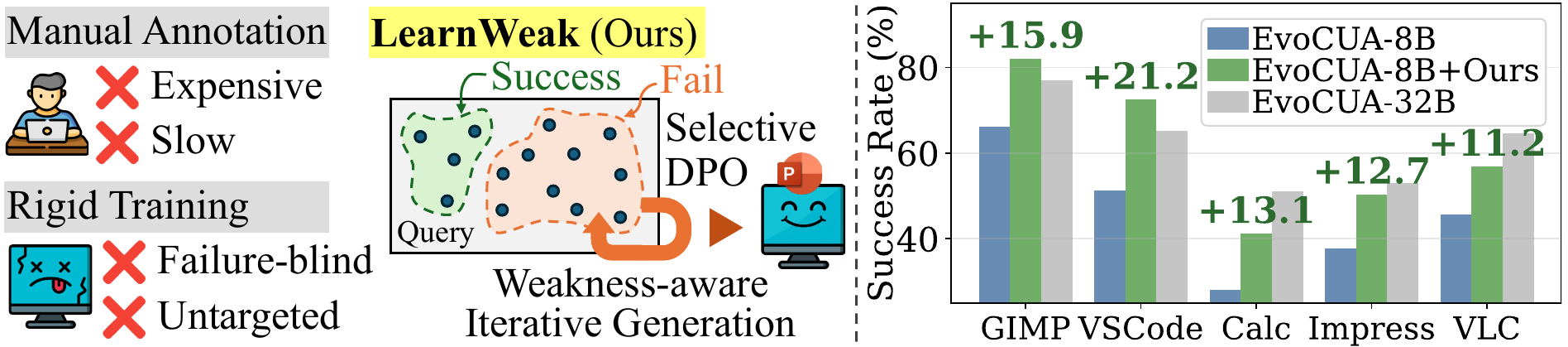}
    \caption{
    Conceptual illustration of \modelname{} and performance gains after domain specialization, showing consistent improvements of the small student across target software domains.
    }
    \vspace{-0.1in}
    \label{fig:concept}
\end{figure}

Domain specialization, which fine-tunes agents on a single target domain, is a promising approach for improving the performance of small CUAs. 
Small models often lack the capacity to simultaneously learn the workflows of diverse software environments, and training across heterogeneous computer-use tasks can lead to catastrophic forgetting and degraded performance within individual domains~\cite{xu2025functionvector, wu2025mitigating, liu2026continual}.
Although scaling up data or designing more sophisticated training objectives can help, both require significant annotation effort or computational cost~\cite{opencua, li2024effects, wang2025ui}.
By contrast, domain-specialized training can improve sample efficiency by focusing on domain-specific interaction patterns rather than broad generalization. 
Recent studies~\cite{sun2025seagent, liu2026osexpert, sun2025coda, chen2025risk} provide empirical evidence supporting the effectiveness of this approach for small CUAs.

Domain specialization for CUAs consists of two stages: dataset generation and agent training.
In the dataset generation stage, collecting human trajectories is costly due to the long-horizon nature of computer-use tasks, which makes autonomous trajectory generation essential~\cite{yang2025zerogui, xie2025agentsynth}.
Existing fixed data generation strategies do not consider student deficiencies on the target domain, resulting in inefficient training~\cite{pcagente, os-genesis, webstar}.
However, data quality matters as much as data quantity: to specialize efficiently, generated queries should target model weaknesses and missing domain knowledge rather than reinforcing already well-learned skills.

In the training stage, domain specialization must preserve pretrained agentic capabilities while selectively repairing weaknesses. 
Small CUAs develop their own reasoning patterns and recovery mechanisms, and naive fine-tuning can distort these by imposing human or large-model reasoning distributions that diverge from the agent's own~\cite{lauffer2025imitation, lyu2025correction}.
Moreover, failure modes are heterogeneous even within a single model: some failures stem from incorrect planning, whereas others arise from execution errors such as inaccurate coordinates~\cite{gou2025uground, xie2024osworld, Agent-S2}.
These challenges call for a framework that identifies the student's weaknesses in the target domain and applies tailored training objectives.

To address these challenges, we introduce \modelname{}, a fully automated domain specialization framework for small CUAs that targets student weaknesses across both dataset generation and agent training. 
For the dataset generation stage, we propose an annotation-free pipeline that expands the training set through repeated cycles of teacher-student comparison, weakness analysis, and query synthesis.  
It requires only a small set of seed queries, yet produces a compact and targeted dataset that addresses the student's deficiencies.
For agent training, we introduce an error-aware preference optimization which adaptively targets task-specific weaknesses. It dynamically adjusts the training objective according to the failure type, distinguishing between planning and execution failures.
Together, our student-aware data generation and training enable small CUAs to close capability gaps on the target domain without human annotation.

We evaluate \modelname{} across 8 OSWorld domains~\cite{xie2024osworld} using EvoCUA-8B~\cite{evocua} and OpenCUA-7B~\cite{opencua} as base students. 
Our domain specialization improves average performance by 11.6 and 11.1 percentage points on EvoCUA-8B and OpenCUA-7B, respectively. Notably, the specialized small agents surpass the teacher on several domains, and our data-generation pipeline achieves the strongest gains among autonomous generation baselines under matched budgets. We further show that error-aware preference optimization outperforms alternative offline training strategies, including SFT and standard DPO variants.
We hope this work serves as a foundation for more efficient and targeted domain specialization of small CUAs and encourages future research toward closing the performance gap between small open models and large proprietary agents.

\section{Preliminaries}
\label{sec:preliminaries}
\subsection{Computer-Use Agent}
A computer-use agent (CUA) is a policy that operates within an interactive software environment by perceiving the screen and issuing actions to complete a given task. Since the current screen alone does not reveal the full environment state, CUA settings are better modeled as a partially observable decision process (POMDP)~\cite{pomdp}. Following common practice~\cite{zhou2024webarenarealisticwebenvironment, evocua, dihan2025weboperatoractionawaretreesearch}, we handle this partial observability by conditioning the policy on the full interaction history.

At each step $t$, the agent receives the current screen as a partial observation of the environment state together with the interaction history
$h_t = (o_1, a_1, o_2, a_2, \dots, o_{t-1}, a_{t-1})$,
which records all previously observed screens $o_{<t}$ and executed actions $a_{<t}$. Conditioned on the current context $c_t = (q, o_t, h_t)$ 
where $q$ is the task instruction, the agent policy $\pi$ produces a structured output,
\begin{equation}
    a_t = \pi(c_t) = (r_t,\, s_t,\, e_t).
\end{equation}
It consists of three components: (i) \emph{internal reasoning} $r_t$, which reflects the agent's analysis of the current state; (ii) an \emph{action description} $s_t$, a natural language description of the intended action; and (iii) \emph{tool execution} $e_t = (f_t, p_t)$, the executable action that directly manipulates the environment, consisting of a function type $f_t$ and its parameters $p_t$ such as \texttt{left\_click(x,y)} or \texttt{type(text)}. The agent repeats this process until the task is complete, producing the full trajectory
\begin{equation}
    \tau = (o_1, a_1, o_2, a_2, \dots, o_H, a_H).
\end{equation}

\subsection{Problem Formulation}
We address \emph{domain specialization}, namely domain-specific finetuning of a broadly capable student policy to a target domain. In the CUA setting, each target domain has its own task distribution, interface conventions, and software-specific interaction patterns.
Let $\mathcal{E}$ be a set of target domains, where each domain $d \in \mathcal{E}$ corresponds to a distinct software application or operating environment. We are given a student policy $\pi^S$ pretrained on a broad collection of GUI tasks, a stronger teacher policy $\pi^T$, a small set of $K$ human-provided seed queries $\mathcal{Q}^d_0 = \{q_1, \dots, q_K\}$, and an executable environment equipped with an automatic verifier $V(q, \tau) \in \{0, 1\}$. No further human annotation is assumed.

Our problem consists of two coupled stages. In the first stage, we autonomously generate a domain-specific training dataset $\mathcal{D}^d$ by expanding the seed queries and collecting trajectories from the teacher policy:
\begin{equation}
    \mathcal{D}^d = \texttt{DataGen}(\pi^S, \pi^T, \mathcal{Q}^d_0, V),
\end{equation}
where \texttt{DataGen} denotes the dataset generation process that produces training samples without human annotation. In the second stage, we use the generated dataset to train a domain-specialized student:
\begin{equation}
    \hat{\pi}^{S,d} = \arg\min_{\pi_\theta}\, \mathcal{L}\big(\pi_\theta;\, \mathcal{D}^d\big).
\end{equation}
The overall objective is to maximize expected task success on the target domain:
\begin{equation}
    \max_{\pi_{\theta}}\,
    \mathbb{E}_{q \sim \mathcal{Q}^d_{\text{eval}}}
    \left[
        V(q,\,\tau({\pi_{\theta}},q))
    \right],
\end{equation}
where $\tau(\pi_\theta, q)$ denotes the trajectory induced by the policy $\pi_\theta$ on task query $q$, and $\mathcal{Q}_{\mathrm{eval}}^d$ denotes the target-domain evaluation task distribution.

\begin{figure*}[t]
    \centering
    \includegraphics[width=\linewidth]{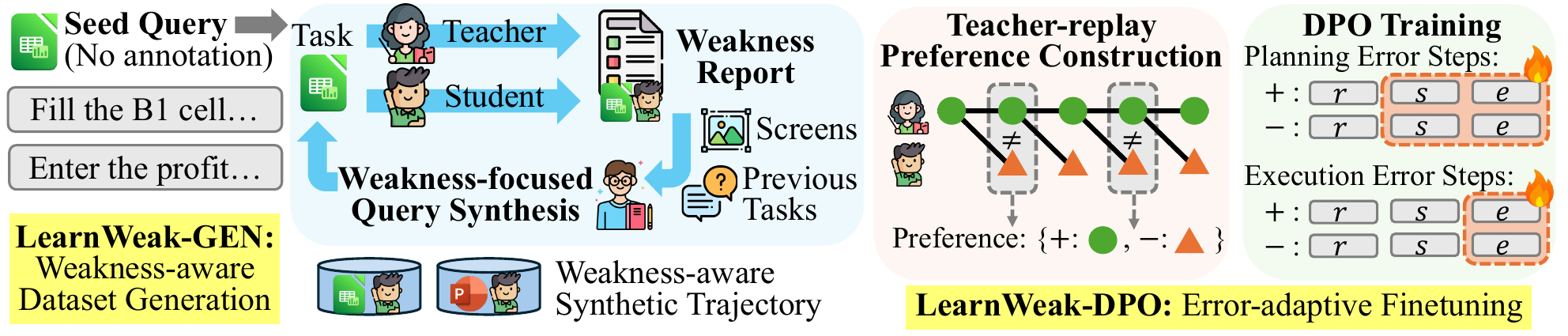}
    \vspace{-0.15in}
    \caption{Overview of \modelname{} framework.
    \modelname{}-GEN iteratively constructs domain data by comparing teacher and student responses, summarizing student weaknesses, and generating new tasks conditioned on weakness reports and representative screenshots. \modelname{}-DPO then converts specializes the student with step-wise preference supervision and error-aware optimization. 
    }
    \vspace{-0.1in}
    \label{fig:method}
\end{figure*}

\section{Method}
\label{sec:method}
\modelname{} decomposes domain specialization into two stages: an annotation-free data generation loop that exposes the current student's domain-specific weaknesses, and the student agent training to correct their behaviors through teacher guidance. 
We first construct the training dataset through iterative teacher-student comparison, verification, and synthetic query generation (\cref{subsec:method_datagen}). 
We then convert the resulting failures into step-wise training signals and specialize the student with domain-specific updates using a selective training objective based on DPO (\cref{subsec:method_train}). 

\subsection{Weakness-Aware Data Generation (\modelname{}-GEN)}
\label{subsec:method_datagen}
We present our annotation-free dataset generation pipeline, which begins with seed query setup, proceeds through iterative cycles of weakness discovery and query synthesis, and concludes with final filtering. 
A formal algorithmic description of our pipeline is provided in \cref{appendix:datagen_algorithm}.

\paragraph{Seed query setup.}
For each target domain $d$, we initialize a small set of executable environment configurations and seed tasks $\mathcal{Q}_0^d$. 
These initial states are constructed separately from the evaluation benchmark so that data generation does not rely on benchmark-specific assets or leaked task states. 
The number of seed queries is small enough that a human can complete the setup within an hour. 

\paragraph{Weakness discovery.}
Weakness discovery is driven by paired teacher-student execution. 
For each task $q \in \mathcal{Q}_i^d$ at iteration $i$, beginning from the seed queries $\mathcal{Q}_0^d$ at $i=0$, we run a teacher trajectory $\tau_q^T$ and a student trajectory $\tau_q^S$ in the same environment, where $\tau_q^S$ is produced by the fixed pre-adaptation snapshot of student $\pi^S$.  
A verifier $V$ is then applied to both trajectories, yielding binary success outcomes $v_q^T, v_q^S \in \{0,1\}$ and structured rationales $r_q^T, r_q^S$.
For student-failure driven generation, we collect the tasks $\mathcal{F}_i^d$ where the teacher is verified to succeed while the student fails:
\begin{equation}
    \begin{aligned}
    (v_q^T, r_q^T) &= V(q, \tau_q^T), \qquad
    (v_q^S, r_q^S) = V(q, \tau_q^S), \\
    \mathcal{F}_i^d
    &=
    \{ q \in \mathcal{Q}_i^d
    \mid
    v_q^T=1,\;
    v_q^S=0
    \}.
    \end{aligned}
\end{equation}
Since the teacher succeeds on these tasks, task infeasibility or invalid environment states are unlikely to be the cause of failure, and student errors can be reliably attributed to the student's own deficiencies.
Finally, the verifier diagnostics from the failure set are summarized into a weakness report $R_i^d$ that captures recurring failure modes in domain $d$, such as incorrect operation selection, inaccurate element localization, or invalid action arguments:
\begin{equation}
    R_i^d = \texttt{Summarize}\bigl(\{ r_q^S \mid q \in \mathcal{F}_i^d \}\bigr).
\end{equation}

\paragraph{Screenshot-guided query generation.}
To generate new queries, we first construct a representative screenshot set $S_i^d$ from both teacher and student trajectories of the current iteration via representation-level clustering and VLM-based reranking, selecting screenshots that are both diverse and semantically informative. 
These screenshots ground the generated queries in realistic environment states, encouraging coverage of diverse software functionalities while reducing the generation of infeasible tasks.
We then employ a task-query generator $G$ to synthesize queries for the next iteration, conditioned on previously generated tasks $\mathcal{Q}_{i}^{d}$, the current weakness report $R_i^d$, the selected screenshots $S_i^d$, and domain-level environment metadata $M^d$ such as available assets. 
Query synthesis proceeds via two complementary strategies: weakness-focused synthesis, which generates tasks conditioned on the weakness report to target identified deficiencies, and exploration-focused synthesis, which omits the report and instead relies on screenshots to generate tasks covering unexplored functionalities or UI elements.   
Using both strategies together maintains a balance between student-aware targeting and open-ended domain exploration:
\begin{equation}
    \begin{gathered}
    \mathcal{Q}_{i+1}^{\text{weak}}
    =
    G(\mathcal{Q}_{i}^{d}, R_i^d, S_i^d, M^d), \qquad
    \mathcal{Q}_{i+1}^{\text{explore}}
    =
    G(\mathcal{Q}_{i}^{d}, \varnothing, S_i^d, M^d), \\
    \mathcal{Q}_{i+1}^{d}
    =
    \mathcal{Q}_{i+1}^{\text{weak}}
    \cup
    \mathcal{Q}_{i+1}^{\text{explore}}.
    \end{gathered}
\end{equation}

\paragraph{Iterative generation.}
Let $N$ denote the total number of generation iterations. We repeat the two stages above, weakness discovery and screenshot-guided query generation, for $N-1$ iterations. 
Each iteration gradually shifts the generated task distribution toward regions that continue to expose unresolved weaknesses, while exploration-focused synthesis maintains diversity in query objectives throughout. 
After all iterations are complete, we aggregate the failed task sets into a final task set:
\begin{equation}
    \mathcal{F}^d(\pi^S) = \bigcup_{i=0}^{N-1}\mathcal{F}_i^d,
\end{equation}
and construct the corresponding teacher-student trajectory collection for the collected tasks:
\begin{equation}
    \mathcal{D}^d(\pi^S) = \{(q,\tau_q^T,\tau_q^S)\mid q \in \mathcal{F}^d(\pi^S)\},
\end{equation}
where $\pi^S$ denotes the fixed student snapshot used for data construction. For brevity, we write $\mathcal{F}^d$ and $\mathcal{D}^d$ omitting the $\pi^S$ dependence in the remainder of this section.

\subsection{Agent Training for Domain Specialization (\modelname{}-DPO)}
\label{subsec:method_train}
We now introduce our CUA training method which adaptively selects the training objective for different failure types while preserving pretrained reasoning capability of the student agent. 
We train the student with DPO~\cite{dpo} on the failure-focused dataset $\mathcal{D}_{\mathrm{pref}}^d$.

\paragraph{Teacher-replay preference construction.}
Trajectory-wise training of CUAs is resource-intensive due to multiple screenshots and long-context reasoning traces, so we intend to apply step-level supervision. 
Even within a failed student trajectory, some steps are already correct. For efficient training, we therefore focus only on steps that require correction, filtering out those where the teacher and student produce the same tool execution.
In detail, for each task $q \in \mathcal{F}^{d}$, we replay the teacher trajectory step by step. 
At each step $t$, we query the student policy $\pi^S$ using a teacher context $c_t^T = (q, o_t^T, h_t^T)$ and obtain a replayed student response $\hat{a}_t^S \sim \pi^S(\cdot \mid c_t^T).$ 

If the action executions of the teacher and the replayed student differ, we build a preference tuple:
\begin{equation}
    (c_t^T, a_t^+, a_t^-)
    =
    (c_t^T, a_t^T, \hat{a}_t^S),
\end{equation}
and aggregate these into a domain-specific preference dataset:
\begin{equation}
    \mathcal{D}_{\mathrm{pref}}^d = \{(c_t^T, a^T_t, \hat{a}_t^S) \mid q \in \mathcal{F}^d,\ t \in \mathcal{T}_q^d \},
\end{equation}
where $\mathcal{T}_q^d = \{t \mid e_t^T \neq \hat{e}_t^S\}$ denotes the set of steps at which the teacher and replayed student produce differing tool executions.
This procedure yields step-level supervision without human annotation, where the teacher trajectory provides a verified successful context and the replayed student response identifies the behavior to be corrected.

\paragraph{Error-aware preference optimization.}
Recall that tool execution $e_t$ is decomposed into a function type $f_t$ and parameters $p_t$, we define a failure type $\epsilon_t$ in two categories: \emph{planning-level error} ($\epsilon_{\text{PLAN}}$) when $f_t^T \ne \hat{f}_t^S$ and \emph{execution-level error} ($\epsilon_{\text{EXEC}}$) when $f_t^T = \hat{f}_t^S$ but $p_t^T \neq \hat{p}_t^S$. 
Let $\pi_\theta$ denote the trainable student policy and $\pi_{\mathrm{ref}}$ the frozen reference policy initialized from the base student. 
Each preference example is associated with a binary mask $m$ over the token position $j$ of $a_t = (r_t, s_t, e_t)$, denoted as:
\begin{equation}
    m^{(j)} =
    \begin{cases}
        0 & \text{if } a_t^{(j)} \in r_t, \\
        g(t) & \text{if } a_t^{(j)} \in s_t, \\
        1 & \text{if } a_t^{(j)} \in e_t,
    \end{cases}
    \qquad
    g(t) =
    \begin{cases}
        1 & \text{if } \epsilon_t =\epsilon_{\text{PLAN}}, \\
        0 & \text{otherwise}.
    \end{cases}
\end{equation}
Since the chosen and rejected responses may have different token lengths, $m$ denotes the response-wise mask instantiated for each score term using the same rule. We define the masked action score as
\begin{equation}
    s_\theta(c, a_t; m)
    =
    \sum_{j=1}^{|a_t|}
    m^{(j)} \log \pi_\theta(a_t^{(j)} \mid c, a_t^{(<j)}),
\end{equation}
and define $s_{\mathrm{ref}}(c,a;m)$ analogously using $\pi_{\mathrm{ref}}$. 
We then optimize
\begin{equation}
    \begin{aligned}
    \mathcal{L}_{\mathrm{DPO}}
    =
    -
    \mathbb{E}_{(c_t,\, a_t^+,\, a_t^-)\sim\mathcal{D}_{\mathrm{pref}}^d}
    \Big[
        \log \sigma \Big(
            \beta \big(
                s_\theta(c_t,a_t^+;m) - s_\theta(c_t,a_t^-;m) \\
                \qquad\qquad
                -\, s_{\mathrm{ref}}(c_t,a_t^+;m) + s_{\mathrm{ref}}(c_t,a_t^-;m)
            \big)
        \Big)
    \Big],
    \end{aligned}
\end{equation}
where $\sigma(\cdot)$ denotes the logistic sigmoid function and $\beta$ is a temperature hyperparameter controlling the strength of the preference signal.
This objective increases the relative likelihood of teacher actions over replayed student actions while restricting updates to the behaviorally relevant span. As a result, the training signal targets the student's actual weakness rather than uniformly relearning the entire action sequence.

\paragraph{Domain scalability.}
Finally, we instantiate each domain specialist through a modular domain-specific update on top of the shared student.
We adopt a modular specialization setting in which domain-specific knowledge is attached to the shared student through domain-dependent updates. 
Specifically, we freeze the student and only update LoRA~\cite{lora} adapters $\{\Delta^d\}_{d \in \mathcal{E}}$. The policy specialized to domain $d$ is written as:
\begin{equation}
    \hat{\pi}^{S,d} = \pi^S \oplus \Delta^d,
\end{equation}
where $\oplus$ denotes attaching the LoRA adapter to the base policy.
At deployment time, the base policy $\pi^S$ is shared across domains, while the adapter corresponding to the current domain is activated to obtain the specialist. This design localizes domain knowledge to domain-specific modules and provides a scalable mechanism for handling multiple domains.

\section{Experiments}
\label{sec:experiments}

\begin{table}[t]
\centering
\setlength{\tabcolsep}{8pt}
{\setlength{\fboxsep}{0.9pt}
\caption{Domain specialization results on OSWorld. Each entry reports mean success rate (\%). Yellow and blue denote the teacher policy and specialized student with \modelname{}, respectively.}
\label{tab:mainresult_osworld}
}
\resizebox{\linewidth}{!}{
\begin{tabular}{l cccccccc >{\columncolor{mygray}}c}

\toprule
~ 
& \textbf{Gimp} 
& \textbf{Calc}
& \textbf{Impress}
& \textbf{Writer}
& \textbf{OS}
& \textbf{Thunderbird} 
& \textbf{VLC}
& \textbf{VSCode} 
& \textbf{Avg.}\\
\midrule
\multicolumn{9}{l}{\textbf{\textit{Generalized Models}}} & \cellcolor{mygray} \\
\hspace{1em}Kimi K2.6~\cite{team2026kimi} & 73.08 & 80.85 & 82.19 & 73.91 & 79.17 & 80.00 & 75.71 & 91.30 & 79.53 \\
\hspace{1em}Claude Sonnet 4.6~\cite{anthropic2026sonnet46} & 69.23 & 74.47 & 70.21 & 86.83 & 91.67 & 66.67 & 81.41 & 72.73 & 76.65 \\
\hspace{1em}Qwen3.5-27B~\cite{team2026qwen3} & 39.74 & 22.70 & 43.97 & 52.17 & 41.67 & 66.67 & 44.12 & 47.83 & 44.86\\
\midrule
\multicolumn{9}{l}{\textbf{\textit{Domain Specialized CUA Models}}} & \cellcolor{mygray} \\
\hspace{1em}SEAgent~\cite{sun2025seagent} & 42.30 & - & 22.70 & 31.80 & - & - & 35.30 & 40.50 & -\\
\hspace{1em}OSExpert~\cite{liu2026osexpert} & 30.80 & 44.70 & 42.60 & 34.70 & - & - & - & - & -\\
\midrule
\multicolumn{9}{l}{\textbf{\textit{CUA Models}}} & \cellcolor{mygray}\\

\rowcolor{myyellow}
\hspace{1em}EvoCUA-32B~\cite{evocua} & 76.29 & 51.06 & 52.98 & 65.22 & 75.00 & 60.00 & 64.65 & 65.22 & 63.80 \\
\hspace{1em}OpenCUA-32B~\cite{opencua} & 74.36 & 35.46 & 48.21 & 56.52 & 61.11 & 57.78 & 37.25 & 72.73 & 55.43 \\

\noalign{\vskip 0.25ex}\cdashline{1-10}\noalign{\vskip 0.75ex}
\hspace{1em}EvoCUA-8B~\cite{evocua} 
& 66.15 & 28.07 & 37.66 & 50.43 & 60.83 & 65.33 & 45.71 & 51.30 & 50.69 \\
\rowcolor{myblue}
\hspace{1em}EvoCUA-8B \textbf{+ Ours} 
& 82.05 & 41.13 & 50.35 & 55.07 & 66.67 & 73.33 & 56.86 & 72.46 &  62.24 \\
\hspace{1em}$\Delta$ & \textbf{+15.9} & \textbf{+13.1} & \textbf{+12.7} & \textbf{+4.6} & \textbf{+5.8} & \textbf{+8.0} & \textbf{+11.2} & \textbf{+21.2} & \textbf{+11.6} \\

\noalign{\vskip 0.25ex}\cdashline{1-10}\noalign{\vskip 0.75ex}
\hspace{1em}OpenCUA-7B~\cite{opencua} 
& 48.46 & 11.91 & 31.49 & 30.43 & 40.00 & 54.67 & 32.94 & 51.30 & 37.65 \\
\rowcolor{myblue}
\hspace{1em}OpenCUA-7B \textbf{+ Ours} 
& 57.69 & 19.15 & 36.88 & 40.58 & 59.42 & 66.67 & 47.06 & 62.32 & 48.72 \\
\hspace{1em}$\Delta$ & \textbf{+9.2} & \textbf{+7.2} & \textbf{+5.4} & \textbf{+10.2} & \textbf{+19.4} & \textbf{+12.0} & \textbf{+14.1} & \textbf{+11.0} & \textbf{+11.1} \\

\bottomrule
\end{tabular}
}
\vspace{-0.15in}
\end{table}

\subsection{Experimental Setup}
\label{sec:experimental_setup}

\paragraph{Benchmarks.}
We employ OSWorld~\cite{xie2024osworld}, a computer-use benchmark covering diverse desktop applications and operating-system utilities. We evaluate our framework on 8 domains: \texttt{Gimp}, \texttt{Libreoffice Calc}, \texttt{Libreoffice Impress}, \texttt{Libreoffice Writer}, \texttt{OS}, \texttt{Thunderbird}, \texttt{VLC}, and \texttt{VSCode}. The entire process, including data generation and training, is performed independently for each domain. During inference, we set the maximum number of steps to 50 for all models and report the average success rate over three trials.

\vspace{-0.1in}
\paragraph{CUA Baselines.} 
To validate the effectiveness of our specialization method, we compare \modelname{} against three categories of systems. First, we include general-purpose frontier and open models, including Claude Sonnet 4.6~\cite{anthropic2026sonnet46}, Kimi K2.6~\cite{team2026kimi}, and Qwen3.5-27B~\cite{team2026qwen3}. Second, we compare with domain-specialized CUA models such as SEAgent~\cite{sun2025seagent} and OSExpert~\cite{liu2026osexpert}. Lastly, we compare against the open CUA families including EvoCUA~\cite{evocua} and OpenCUA~\cite{opencua}. 

\vspace{-0.1in}
\paragraph{Data-generation Baselines.}
To validate that weakness-focused generated data is useful for training the student model, we compare \modelname{} against an existing dataset and other data-construction baselines for CUAs.
First, we compare against a supervision setting based on the AgentNet~\cite{opencua} dataset, which contains a large number of human-validated trajectories. We consider two variants: one that uses all trajectories in AgentNet, and another that samples $N$ trajectories to match the training budget of the other baselines.
Second, we compare with a minimally annotated synthesis pipeline, Trajectory Boosting~\cite{pcagente}, which expands a small set of human trajectories by generating possible action spaces. 
Lastly, we compare with zero-human annotation generators such as AgentSynth~\cite{xie2025agentsynth}, OS-Genesis~\cite{os-genesis}, and ZeroGUI~\cite{yang2025zerogui}. 
Additionally, we apply WebSTAR~\cite{webstar}, a step-level filtering method that selects useful training steps from existing trajectories, to our generated data and report the results. All methods are evaluated under the same setting including student backbone and specialization budget such as dataset amount or training time.

\vspace{-0.1in}
\paragraph{Implementation Details.} 
We experiment on EvoCUA-8B and OpenCUA-7B as the student models to be specialized, and EvoCUA-32B as the teacher policy for data construction. 
Unless otherwise specified, all subsequent analyses use EvoCUA-8B as the student model.
We provide additional details, including hyperparameters and training budget, in \cref{appendix:implementation_details}, and the prompt templates used for our dataset-generation mechanism in \cref{appendix:prompt_templates}.

\subsection{Domain Specialization Results}
\cref{tab:mainresult_osworld} shows that \modelname{} yields consistent improvements for both small CUA backbones across all eight OSWorld domains. Averaged over domains, our specialization improves EvoCUA-8B from 50.69 to 62.24 and OpenCUA-7B from 37.65 to 48.72, corresponding to gains of 11.6 and 11.1 percentage points, respectively. The improvements are not confined to a single application type, but are observed across office software, system utilities, visual editing, and coding-oriented workflows.

\textbf{\textit{Weakness-focused specialization enables small student to surpass the teacher in several domains.}} Our specialized EvoCUA-8B model outperforms the 32B teacher on \texttt{Gimp}, \texttt{Thunderbird}, and \texttt{VSCode}. This suggests that weakness-focused corrective supervision can be more than simple imitation: even when the training data is conditioned by the teacher, the student can use corrections to address its own domain-specific failures and surpass the teacher in selected domains. 

\textbf{\textit{Specialization gains arise from different domains depending on the student model.}} For EvoCUA-8B, the largest improvements appear in \texttt{VSCode}, \texttt{Gimp}, \texttt{Calc}, and \texttt{Impress}, whereas for OpenCUA-7B the strongest gains appear in \texttt{OS}, \texttt{VLC}, \texttt{Thunderbird}, and \texttt{VSCode}. This variability suggests that specialization depends less on domain difficulty alone and more on how well each student model adapts to the interaction patterns of a given software domain.

\begin{table}[!t]
    \centering
    \begin{minipage}[c]{0.54\linewidth}
        \caption{Comparison with data-construction baselines on the four OSWorld domains. 
        We report mean success rate (\%) under a matched training budget.}
        \label{tab:datagen_baselines}
        \centering
\renewcommand{\arraystretch}{1.1}
\resizebox{\linewidth}{!}{
\begin{tabular}{l cccc>{\columncolor{mygray}}c}
\toprule
~
& \textbf{Calc}
& \textbf{Impress}
& \textbf{VLC}
& \textbf{VSCode} 
& \textbf{Avg.} \\
\midrule
Zero-shot & 28.07 & 37.66 & 45.71 & 51.3 & 40.69 \\
\midrule
\multicolumn{5}{l}{\textbf{\textit{Existing Data}}} & \cellcolor{mygray} \\
\hspace{1em}AgentNet~\cite{opencua} & 34.04 & 39.01 & 49.01 & 69.57 & 47.91  \\
\hspace{1em}AgentNet {\small (N-sampled)} & 32.62 & \underline{40.43} & 49.02 & 63.77 & 46.46  \\
\midrule
\multicolumn{5}{l}{\textbf{\textit{Minimal Human Annotation}}} & \cellcolor{mygray} \\
\hspace{1em}Trajectory Boosting~\cite{pcagente} & 30.50 & 19.88 & 45.10 & 49.28 & 36.19  \\
\midrule
\multicolumn{5}{l}{\textbf{\textit{Zero Human Annotation}}} & \cellcolor{mygray} \\
\hspace{1em}AgentSynth~\cite{xie2025agentsynth} & 31.21 & 39.01 & 39.22 & 71.01 & 45.11  \\
\hspace{1em}OS-Genesis~\cite{os-genesis} & 31.91 & 37.59 & 45.10 & 68.12 & 45.68  \\
\hspace{1em}ZeroGUI~\cite{yang2025zerogui} & \underline{36.17} & \underline{40.43}	& 48.86	& 62.30 & 46.94 \\
\hspace{1em}WebSTAR~\cite{webstar} & 31.21 & \underline{40.43} & \underline{52.94} & \textbf{73.91} & \underline{49.62}  \\
\rowcolor{myblue}
\hspace{1em}\textbf{\modelname{}} & \textbf{41.13} & \textbf{50.35} & \textbf{56.86} & \underline{72.46} & \textbf{55.20}  \\
\bottomrule
\end{tabular}
}
   
    \end{minipage} 
    \hfill
    \begin{minipage}[c]{0.45\linewidth}
        \caption{Effect of the weakness-report source model in \modelname{}-GEN. We specialize the base model $\pi_{\theta}$ using datasets constructed from weakness reports derived from different source students $\pi^{S}$.}
        \label{tab:abl_nontarget}
        \centering
\resizebox{\linewidth}{!}{
\begin{tabular}{l cccc}
\toprule
\multirow{2}{*}{\makecell{$\pi_{\theta}$ / \textbf{$\pi^{S}$}}}
& \multirow{2}{*}{\textbf{Calc}} 
& \multirow{2}{*}{\textbf{Impress}}
& \multirow{2}{*}{\textbf{VLC}}
& \multirow{2}{*}{\textbf{VSCode}}\\
& & & & \\
\midrule
\multicolumn{5}{l}{\textbf{\textit{OpenCUA-7B}}} \\
\hspace{1em}Zero-shot & 11.91 & 31.49 & 32.94 & 51.30 \\
\hspace{1em}EvoCUA-8B & 9.93 & 27.54 & 45.10  & 50.72 \\
\hspace{1em}UI-TARS-1.5-7B & 7.80 & 33.35 & - & 49.28 \\
\cellcolor{myblue}\hspace{1em}OpenCUA-7B & \cellcolor{myblue}\textbf{19.15} & \cellcolor{myblue}\textbf{36.88} & \cellcolor{myblue}\textbf{47.06} &\cellcolor{myblue}\textbf{62.32} \\
\midrule
\multicolumn{5}{l}{\textbf{\textit{EvoCUA-8B}}} \\
\hspace{1em}Zero-shot & 28.07 & 37.66 & 45.71 & 51.30 \\
\hspace{1em}UI-TARS-1.5-7B & 22.70 & 31.21 & - & 71.01 \\
\hspace{1em}OpenCUA-7B & 39.01 & 43.26 & 47.06 & \textbf{73.91} \\
\cellcolor{myblue}\hspace{1em}EvoCUA-8B & \cellcolor{myblue}\textbf{41.13} & \cellcolor{myblue}\textbf{50.35} & \cellcolor{myblue}\textbf{56.86} &\cellcolor{myblue}72.46 \\
\bottomrule
\end{tabular}
}

    \end{minipage}
\vspace{-0.2in}
\end{table}

\vspace{-0.1in}
\subsection{Comparison with Dataset Construction Baselines}
\label{sec:dataset_effective}

In \cref{tab:datagen_baselines}, we compare \modelname{}-GEN against alternative data construction pipelines under a matched training budget: existing human-validated data, minimal human annotation, and zero human annotation.
First, fine-tuning on existing AgentNet trajectories yields only limited gains, even when using the full set of human-validated trajectories, suggesting that simply reusing existing supervision is insufficient for effective domain specialization. 
Second, the minimal human annotation baseline, Trajectory Boosting, further degrades performance, indicating that expanding the action space around fixed states does not provide useful supervision without sufficient exploration of domain-relevant states.
Lastly, zero human annotation setting such as AgentSynth, OS-Genesis, and ZeroGUI perform comparably to AgentNet retraining. Although these methods explore the computer-use environments and use LLMs or VLMs to generate tasks, their generation process is weakness-agnostic, as it does not account for the target model's observed failure modes.

\modelname{} achieves the best average performance, outperforming WebSTAR by 5.58 percentage points.
Since WebSTAR contributes a data-filtering strategy rather than a generation pipeline, we apply its filtering criterion to the same weakness-aware dataset produced by our generation procedure, therefore WebSTAR and \modelname{} differ only in the filtering stage. However, it remains weakness-agnostic, scoring each step by generic quality rather than by the target model’s observed failures, whereas \modelname{} retains trajectories aligned with the student’s identified weaknesses.
We also find that improvements are not uniform across domains. In the \texttt{VSCode} domain, all methods reach comparable performance, leaving little room for weakness-aware specialization to provide additional gains. The advantage of \modelname{} is instead most evident in the remaining domains, where it outperforms every baseline that explores without targeting the model's weaknesses.
\section{Analysis}
\label{sec:analysis}
\subsection{Data Generation Pipeline Analysis}
\label{subsec:abl_datagen_pipeline}
\vspace{-0.05in}

\paragraph{Weakness-awareness.}
To verify that our dataset generation captures model-specific weaknesses, we train each target model ($\pi_\theta$) on datasets constructed from weakness reports derived from different source students ($\pi^S$) , as shown in \cref{tab:abl_nontarget}. Because failure cases and weakness types differ across student models, a student-aware generator  produce the most useful data when the weakness report is derived from the target model itself.
Both OpenCUA-7B and EvoCUA-8B achieve the highest performance when trained on datasets generated from their own failure cases, while cross-student datasets yield consistently lower gains.
This confirms that our weakness-focused generation can focus the most useful data distribution, validating the key nature of \modelname{}-GEN.

\vspace{-0.05in}
\paragraph{Pipeline Components.}
We conduct an ablation study on the key modules of \modelname{}-GEN: (i) iterative generation, by comparing against a one-shot generation variant that produces the same number of trajectories in a single pass without either iteration or weakness-report conditioning; and (ii) weakness-report conditioning itself.
\cref{tab:abl_datagen} shows that domain specialization without iterative generation (one-shot generation) can already be useful relative to the zero-shot student, improving the average score from 40.69 to 48.82. 
However, adding iterative generation without weakness-report conditioning does not improve upon the one-shot generation, demonstrating that exploration-only generation fails to collect effectively targeted training samples.
By contrast, the full pipeline, which combines iterative generation with weakness-report conditioning, achieves the best average result and the strongest performance on three of the four domains. 
These results suggest that the benefit of iterative expansion depends on student-aware guidance: repeated generation becomes most effective when it is steered by the student's observed failure patterns rather than by domain exploration alone.

\vspace{-0.05in}
\paragraph{Teacher Choice.}
\label{par:teacher_choice}
\begin{table}[!t]
    \centering
    \begin{minipage}[c]{0.45\linewidth}
        \centering
\setlength{\tabcolsep}{3pt}
\renewcommand{\arraystretch}{1.1}
\captionof{table}{Ablation on teacher policy ($\pi_T$) choice for data generation. We report the performance of the teachers and the corresponding teacher-guided specialized students.
}
\vspace{0.05in}
\label{tab:abl_teacher}
\resizebox{\linewidth}{!}{
\begin{tabular}{l cc cc}
\toprule
\multirow{2}{*}{\makecell{Teacher Policy}}
& \multicolumn{2}{c}{Teacher}
& \multicolumn{2}{c}{Specialized Student} \\
\cmidrule(lr){2-3} \cmidrule(lr){4-5}
& \textbf{Calc} & \textbf{VSCode} 
& \textbf{Calc} & \textbf{VSCode}  \\
\midrule
Zero-shot 
& -- & -- 
& 28.07 & 51.30  \\
Claude Haiku 4.6 
& 36.17 & 69.60 
& 30.50 & 71.01 \\
EvoCUA-32B 
& 51.06 & 65.22 
& \textbf{41.13} & 72.46  \\
Kimi K2.5
& 63.83 & 86.96
& \textbf{41.13} & \textbf{73.91}  \\
\bottomrule
\end{tabular}
}
   
    \end{minipage}
    \hfill
    \begin{minipage}[c]{0.52\linewidth}
        \caption{Comparison on training objectives. For {\footnotesize \modelname{}-SFT}, we adapt ours into SFT objective.}
        \label{tab:abl_trainmethod}
        \centering
\setlength{\tabcolsep}{5pt}
\resizebox{\linewidth}{!}{
\begin{tabular}{l cccc >{\columncolor{mygray}}c}
\toprule
 ~ 
& \textbf{Calc}
& \textbf{Impress}
& \textbf{VLC}
& \textbf{VSCode} 
& \textbf{Avg.} \\
\midrule
Zero-shot & 28.07 & 37.66 & 45.71 & 51.30 & 40.69 \\

\midrule

\multicolumn{2}{l}{\textbf{\textit{SFT}}} \\
No masking (standard) & 29.08 & 39.72 & 45.10 & 68.12 & 45.51 \\
\modelname{}-SFT & \underline{34.04} & \underline{46.81} & 45.10 & 69.57 & \underline{48.88} \\

\midrule
\multicolumn{2}{l}{\textbf{\textit{DPO}}} \\
No masking (standard)  & 27.66 & 40.43 & \underline{49.02} & 65.22 & 45.58 \\
$m_j = 1$ if $a_t^{(j)} \in \{r_t\}$ & 18.44 & 17.02 & 41.18 & 63.77 & 35.10 \\
$m_j = 1$ if $a_t^{(j)} \in \{r_t, s_t\}$ & 24.82 & 39.72 & 45.10 & \underline{71.01} & 45.16 \\
\modelname{}-DPO & \textbf{41.13} & \textbf{50.35} & \textbf{56.86} & \textbf{72.46} & \textbf{55.20} \\
\bottomrule
\end{tabular}
\vspace{-0.1in}
}
   
    \end{minipage}
\vspace{-0.1in}
\end{table}

\cref{tab:abl_teacher} evaluates how teacher choice ($\pi^{T}$) affects specialization. In this ablation, we compare Claude Haiku 4.6, EvoCUA-32B, and Kimi K2.5. The main pattern is that teacher strength matters up to a point: the weaker teacher, Claude Haiku 4.6, yields smaller gains than others on both domains. At the same time, the stronger two teachers EvoCUA-32B and Kimi K2.5 produce very similar specialized student performances, despite a large gap in their own standalone success rates. This suggests that teacher capability matters mainly because it helps generate reliable successful trajectories to detect student's weaknesses. Once the teacher is strong enough, further gains depend less on how often the teacher succeeds and more on whether its supervision targets weaknesses that are actionable for the student.

\begin{table}[!t]
    \centering
    \begin{minipage}[c]{0.53\linewidth}
    \captionof{table}{Ablation of the data-generation pipeline design in \modelname{}-GEN.}
    \label{tab:abl_datagen}
    \setlength{\tabcolsep}{2pt}
    \resizebox{\linewidth}{!}{
    \begin{tabular}{ccc cccc >{\columncolor{mygray}}c}
    \toprule
    \makecell{Domain\\Special.}
    & \makecell{Iter. \\ Gen.}
    & \makecell{Weak.\\Report}
    & \textbf{Calc}
    & \textbf{Impress}
    & \textbf{VLC}
    & \textbf{VSCode}
    & \textbf{Avg.}\\
    \midrule
    \xmark & \xmark & \xmark & 28.07 & 37.66 & 45.71 & 51.30 & 40.69 \\
    \cmark & \xmark & \xmark & \underline{34.57} & 39.72 & \underline{47.06} & \textbf{73.91} & \underline{48.82}   \\
    \cmark & \cmark & \xmark & 24.82 & \underline{42.55} & 43.14 & \underline{72.46} & 45.74  \\
    \cmark & \cmark & \cmark & \textbf{41.13} & \textbf{50.35} & \textbf{56.86} & \underline{72.46} & \textbf{55.20} \\
    \bottomrule
    \end{tabular}
    }
    \end{minipage} 
    \hfill
    \begin{minipage}[c]{0.44\linewidth}
    \includegraphics[width=\linewidth]{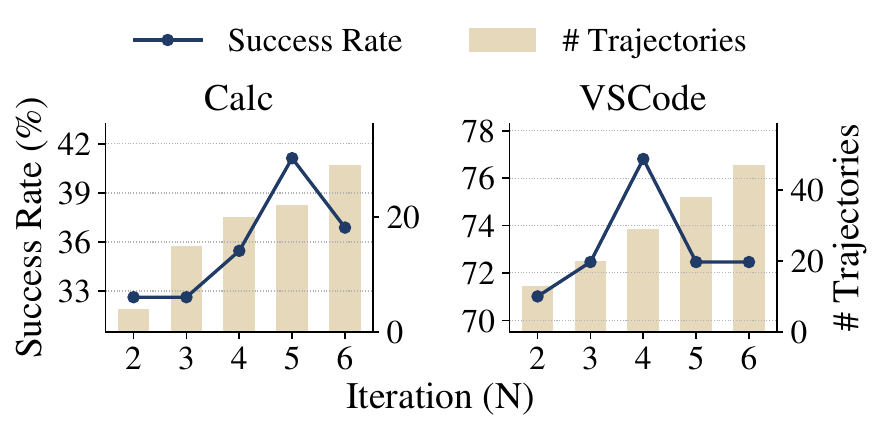}
    \vspace{-0.2in}
    \captionof{figure}{The number of generation iters.}
    \label{fig:abl_iteration}
    \end{minipage}
    \vspace{-0.15in}
\end{table}

\vspace{-0.05in}
\paragraph{The number of Generation Iteration.}
\cref{fig:abl_iteration} indicates that the effect of increasing the number of generation rounds ($N$) is non-monotonic. In both \texttt{Calc} and \texttt{VSCode}, performance improves over the early iterations, reaches a maximum at an intermediate stage, and then decreases as additional rounds are added. 
This pattern suggests that the effectiveness of iterative generation is not determined by data volume alone. Instead, the marginal value of additional rounds appears to depend on whether the newly generated tasks remain well aligned with the student's unresolved weaknesses.

\subsection{Training Objective Analysis}
In \cref{tab:abl_trainmethod}, we compare \modelname{}-DPO with standard SFT, DPO and the variants that follow the defined error types through different supervision scopes using the same generated dataset $\mathcal{D}^d$ from \modelname{}-GEN.
As discussed in \cref{subsec:method_train}, we adaptively train using the error types from the student model's response when it differs from the teacher's. The result shows that full-response optimization is not sufficient for specialization: standard SFT and DPO improve only modestly over the zero-shot student. Error-aware masking consistently improves SFT, and the best results are obtained by \modelname{}-DPO, which outperforms standard DPO by 9.62 points on average. Masking only planning-level or only execution-level tokens is not enough, indicating that effective specialization requires both preference learning and selective updates over both error types.

\section{Related Work}
\label{sec:relatedworks}

\paragraph{Computer-Use Agents (CUAs).}

CUAs complete user-specified tasks by interacting with GUIs through low-level actions such as clicking, typing, and scrolling. Recent advances in VLMs have enabled screenshot-conditioned agents that operate directly in computer environments, with proprietary systems such as Claude Sonnet~\cite{anthropic2026sonnet46} and Kimi~\cite{team2026kimi} demonstrating strong agentic capabilities, and open models such as UI-TARS~\cite{uitarsone}, OpenCUA~\cite{opencua}, and EvoCUA~\cite{evocua} advancing end-to-end vision-language-action modeling. However, execution-based benchmarks~\cite{xie2024osworld, windowsagentarena, yang2025macosworld} reveal persistent domain-dependent performance gaps, particularly in productivity software where application-specific interaction knowledge is required beyond generic UI grounding~\cite{zhao2026worldguiinteractivebenchmarkdesktop, ofengenden2025pptarenabenchmarkagenticpowerpoint}. These gaps motivate domain-specialized small CUAs, which reduce serving cost and latency while focusing capacity on narrower interaction distributions, making them especially attractive for long-horizon tasks in target software domains. Related efforts such as SEAgent~\cite{sun2025seagent} and Fara-7B~\cite{awadallah2025fara7befficientagenticmodel} demonstrate the promise of software-specific adaptation, and our work follows this direction with a focus on sample-efficient specialization without human annotation.

\paragraph{Automated Trajectory Generation.}
Since human-annotated GUI trajectories are expensive to collect, automated trajectory generation is increasingly important.
PC-Agent-E~\cite{pcagente} reduces annotation cost by expanding a small set of human trajectories using stronger models, while recent work explores fully zero-annotation pipelines.
AgentSynth~\cite{xie2025agentsynth} composes successful subtasks into longer-horizon tasks, OS-Genesis~\cite{os-genesis} retrospectively synthesizes task descriptions from environment exploration, ZeroGUI~\cite{yang2025zerogui} combines VLM-based task generation with annotation-free reward estimation, AgentTrek~\cite{xu2025agenttrekagenttrajectorysynthesis} converts web tutorials into executable GUI tasks verified by a VLM evaluator, and Watch-and-Learn~\cite{song2025watch} generates trajectories by grounding instructional videos into executable GUI actions.
These methods scale up trajectory synthesis without human annotation, but primarily target dataset volume or diversity rather than what to generate based on the current model's failures.
In contrast, \modelname{} performs student-aware generation by identifying capability gaps from failed executions and synthesizing tasks conditioned on those weaknesses, making each sample more informative for domain specialization.

\paragraph{Agent Training.}

CUA training is closely related to imitation learning and preference optimization for long-horizon interactive agents. Supervised imitation is a natural baseline for learning GUI action sequences, but it suffers from covariate shift when the policy deviates from expert trajectories and accumulates errors over time. DAgger~\cite{dagger} addresses this by collecting expert labels on learner-induced states, and recent On-Policy Expert Corrections~\cite{lauffer2025imitation} apply a similar idea to multi-turn LM agents. Recent work also uses failures as preference signals. ETO~\cite{song2024trialerrorexplorationbasedtrajectory} constructs contrastive pairs from successful and failed trajectories, while DPO~\cite{dpo} enables direct optimization over such pairwise preferences. Our work focuses on constructing targeted preference data: given a successful teacher trajectory, we sample student rollouts under the same context and form DPO pairs between teacher and student action spans. We further apply error-aware span selection to train only on the segment where the student diverges, making supervision more focused than imitating full trajectories.
\section{Conclusion}
\label{sec:conclusion}
We study domain specialization for small computer-use agents in a fully automated setting. The central idea of \modelname{} is that, for specialization, the most useful supervision is not broad domain coverage but the subset of tasks that exposes the current student's actual weaknesses. Based on this view, we propose a two-stage framework consisting of \modelname{}-GEN, which iteratively constructs a weakness-aware domain dataset through teacher-student comparison and screenshot-grounded query synthesis, and \modelname{}-DPO, which converts the resulting cases, where the teacher succeeds but the student fails, into step-level preference supervision with error-aware masking.
Results show that this targeted specialization strategy substantially improves small CUAs across diverse software domains and outperforms alternative data-construction methods. The gains support our claim that efficient specialization depends on identifying and repairing student-specific weaknesses rather than simply scaling synthetic data. Our results further suggest that automated domain specialization can narrow the gap between small open CUAs and much larger agents without requiring human trajectory annotation, making small specialized agents a more practical deployment path for real-world software environments. Beyond per-domain specialization, our modular LoRA-based design naturally extends to a multi-application deployment scenario in which a library of per-domain adapters is maintained and the adapter matching the target application is activated at inference time. A systematic empirical study of such multi-adapter routing across many domains is a promising direction for future work. 
\section*{Limitation}
\label{appendix:limitation}

Our study has several limitations. First, we assume the availability of a teacher model that provides reasonably reliable guidance within the target domain. If the teacher is highly unstable or systematically biased for a given domain, the resulting supervision may inherit such errors. However, this limitation is not specific to our specialization framework; it reflects a general dependency of teacher-guided offline learning methods on the quality of the supervision source.
Second, our specialization framework focuses on domain knowledge and therefore assumes a student base model that already possesses general computer-use skills at least, including visual grounding, action generation, and error recovery. Therefore, it does not guarantee effective improvement for general-purpose models that have not been trained for computer-use tasks, or for students whose failures primarily arise from missing foundational GUI capabilities rather than domain-specific weaknesses.

\bibliography{main}


\newpage
\appendix
\label{sec:appendix}

\section*{Appendix Overview}

This appendix provides supplementary material for the main paper.

\begin{itemize}[leftmargin=2em, itemsep=2pt, parsep=0pt, topsep=2pt]
    \item Algorithmic details for the data generation pipeline
    \item Implementation details
    \item Additional experimental results and analyses \\ 
    : Generated data statistics, Failure-focused trajectory selection, Other specialization results
    \item Prompt templates
    \item Qualitative results: Weakness reports, Synthetic queries, Case studies
\end{itemize}

\section{Algorithmic Details}
\subsection{Data Generation Pipeline}
\label{appendix:datagen_algorithm}

\cref{alg:datagen} formalizes the per-iteration operation of \modelname{}-GEN described in \cref{subsec:method_datagen}.

\begin{algorithm}[h]
\small
\caption{Data Generation Pipeline}
\label{alg:datagen}
\begin{algorithmic}[1]
\Require Target domain $d$, teacher policy $\pi^T$, fixed student policy $\pi^S$, verifier $V$, task generator $G$, VLM-based screenshot selector $\textsc{Select}$, domain-level metadata $M^d$, number of iterations $N$
\Ensure Aggregated failure task set $\mathcal{F}^d(\pi^S)$ and teacher-student trajectory collection $\mathcal{D}^d(\pi^S)$

\State Initialize seed task set $\mathcal{Q}_0^d$ and environment configurations for domain $d$
\State $\mathcal{D}_{\mathrm{raw}}^d \leftarrow \emptyset$

\For{$i = 0, 1, \ldots, N-1$}
    \Comment{\textit{Weakness Discovery}}
    \For{each $q \in \mathcal{Q}_i^d$}
        \State  $\tau_q^T \leftarrow \textsc{Run}(\pi^T, q)$ \Comment{Teacher trajectory}
        \State Student trajectory $\tau_q^S \leftarrow \textsc{Run}(\pi^S, q)$ \Comment{Student trajectory}
        \State $(v_q^T, r_q^T) \leftarrow V(q, \tau_q^T)$ 
        \Comment{Evaluate teacher trajectory}
        \State $(v_q^S, r_q^S) \leftarrow V(q, \tau_q^S)$ \Comment{Evaluate student trajectory}
        
    \EndFor
    \State $\mathcal{F}_i^d = \{ q \in \mathcal{Q}_i^d \mid v_q^T = 1,\; v_q^S = 0 \}$
    \Comment{Identify failure set}
    \State $R_i^d \leftarrow \textsc{Summarize}(\{ r_q^S \mid q \in \mathcal{F}_i^d \})$
    \Comment{Summarize weakness report}
    \State $\mathcal{D}_{\mathrm{raw}}^d \leftarrow \mathcal{D}_{\mathrm{raw}}^d \cup \{ (q, \tau_q^T, \tau_q^S) \mid q \in \mathcal{F}_i^d \}$

    \If{$i < N - 1$} 
        \State $S_i^d \leftarrow \textsc{Select}(\{ \tau_q^T, \tau_q^S \mid q \in \mathcal{Q}_i^d \})$
        \Comment{Collect representative screenshot set}
        \State $\mathcal{Q}_{i+1}^{\text{weak}} \leftarrow G(\mathcal{Q}_{i}^d,\; R_i^d,\; S_i^d,\; M^d)$ \Comment{Weakness-conditioned}
        \State $\mathcal{Q}_{i+1}^{\text{explore}} \leftarrow G(\mathcal{Q}_{i}^d,\; \varnothing,\; S_i^d,\; M^d)$ \Comment{Unconstrained}
        \State $\mathcal{Q}_{i+1}^d \leftarrow \mathcal{Q}_{i+1}^{\text{weak}} \cup \mathcal{Q}_{i+1}^{\text{explore}}$
    \EndIf
\EndFor
\State $\mathcal{F}^d(\pi^S) \leftarrow \bigcup_{i=0}^{N-1}\mathcal{F}_i^d$
\State $\mathcal{D}^d(\pi^S) \leftarrow \{(q,\tau_q^T,\tau_q^S)\mid q \in \mathcal{F}^d(\pi^S)\}$
\State \Return $\mathcal{F}^d(\pi^S), \mathcal{D}^d(\pi^S)$
\end{algorithmic}
\end{algorithm}

\section{Implementation Details}
\label{appendix:implementation_details}

\subsection{Shared Setup}
\label{appendix:shared_setup}

\paragraph{Benchmark-disjoint Configurations.}
Before generating domain-specific dataset, we first set custom configurations to avoid contamination from benchmark-specific assets.
Many exploration-based generators, including ZeroGUI~\cite{yang2025zerogui} and OS-Genesis~\cite{os-genesis}, operate directly on OSWorld configurations that contain benchmark files such as presentation decks and application-specific documents. In contrast, we construct separate training configurations that are disjoint from the original OSWorld evaluation setups, preventing generated screenshots, interaction traces, and trial-and-error patterns from leaking benchmark-specific artifacts into training. 

For each target domain, we construct 6 environment configurations and 10 seed queries built upon them. Each configuration covers launching the target software and, when applicable, downloading or loading the necessary files. These configurations are designed to be structurally similar to the original OSWorld setups while containing different files and assets as follows:
\vspace{-0.05in}
\begin{itemize}
\setlength{\itemsep}{0pt}
\setlength{\parskip}{0pt}
\item \texttt{GIMP}: image files.
\item \texttt{Libreoffice Calc}: spreadsheet files along with per-sheet data.
\item \texttt{Libreoffice Impress}: presentation files along with per-slide text content.
\item \texttt{Libreoffice Writer}: document files along with their textual content.
\item \texttt{OS}: linux commands (e.g., \texttt{\small mkdir -p /home/user/Project/Project1}).
\item \texttt{Thunderbird}: downloading and setting up an email profile.
\item \texttt{VLC}: video files.
\item \texttt{VS Code}: cloning source code from GitHub.
\end{itemize}
\vspace{-0.05in}
Based on these configurations, we manually authored 10 simple seed queries per domain. This process takes less than two hours of human effort and would be unnecessary in unrestricted software environments, without the constraints of the current docker-based setup. We will release all configurations and seed queries to support reproducibility.

\paragraph{Evaluation.}
All evaluations are conducted in the local docker provider environment in OSWorld. We exclude the \texttt{Chrome} domain from the current evaluation suite because it exhibited weaker reproducibility and less stable evaluation behavior. For each model-domain pair, we run evaluation three times and report the mean success rate. 

\paragraph{Training.}
All experiments are conducted on a single H200 GPU. LoRA fine-tuning for domain specialization on 7--8B models takes under 5 hours, depending on data size. We freeze the vision tower and train LoRA adapters with rank 32 and $\alpha=64$. We use a visual budget of up to $10^6$ image pixels, an effective batch size of 64, a learning rate of $1\times 10^{-6}$, cosine scheduling with 10\% warmup, and train for 20 epochs.
\subsection{\modelname{}}

\paragraph{Data Generation.}
Unless otherwise stated, our main data-generation runs use EvoCUA-8B as the student, EvoCUA-32B as the teacher, and GPT-5-mini~\cite{openai2026gpt54miniNano} for verification, weakness summarization, screenshot ranking, and query generation. Each domain uses 10 seed queries defined over 6 benchmark-disjoint configurations, and we run a total of $N=5$ iterations per domain.
For screenshot selection, we reduce the raw screenshot pool with CLIP-based diversity filtering and then keep the top 10 screenshots after GPT-5-mini ranking. For query generation, we issue 2 calls per configuration, with and without the weakness report, and request 3 instructions per call. This yields up to 36 candidate queries per round and up to 144 generated candidate queries per domain across the 4 iterative rounds after seeding. The generation prompts additionally constrain instructions to be short, executable, and compliant with the workspace and path constraints of the current configuration.

\paragraph{DPO Training.}
To build the step-wise preference dataset, we parse teacher and replayed student outputs into structured tool calls and compare them after removing \texttt{wait} actions. Exact matches are discarded, as are steps that differ only by \texttt{wait}. Usable mismatches are split into two cases: parameter differences, mapped to execution-level errors, and action-type or tool-count differences, mapped to planning-level errors. For coordinate-based actions, teacher and student selections are treated as equivalent when they fall within a 20-pixel tolerance. We then train on this dataset with the DPO loss using $\beta=0.1$.
\subsection{Data-Construction Baselines}

All baseline data-construction methods are re-implemented with our benchmark-disjoint training configurations for each target domain. In this comparison, we target EvoCUA-8B for domain specialization. We use GPT-5-mini~\cite{openai2026gpt54miniNano} as the auxiliary VLM for both verification and query generation across all methods. Unless otherwise stated, we match the per-domain training budget to that of our method, using 22, 52, 32, and 38 trajectories for \texttt{Calc}, \texttt{Impress}, \texttt{VLC}, and \texttt{VS Code}, respectively.

\vspace{-0.05in}
\paragraph{Trajectory Boosting.}
We adopt the trajectory boosting mechanism introduced in PC-Agent-E~\cite{pcagente}. The original procedure constructs training data from a small set of human-annotated trajectories by generating candidate actions for each state. In our implementation, we replace the human-annotated source with 10 teacher trajectories from EvoCUA-32B on our seed queries, and boost them at the step level by generating $\times 8$ candidate actions per state.

\vspace{-0.05in}
\paragraph{OS-Genesis~\cite{os-genesis}.}
We follow the original four-stage pipeline: environment exploration, reverse task synthesis, clean trajectory recollection, and TRM-based filtering. Exploration and recollection are both executed by EvoCUA-32B. For each domain, we first generate a $2\times$ exploration buffer relative to the final target count, synthesize instructions from the resulting exploration trajectories, and then re-execute the synthesized instructions to collect clean trajectories. We keep only trajectories with score at least 3 under the TRM-style evaluator and retain the top-scoring examples under the matched per-domain budget. As in the official implementation, we generate both planning and action data for each retained example and train on them jointly in a single supervised stage.

\vspace{-0.05in}
\paragraph{AgentSynth~\cite{xie2025agentsynth}.}
We follow the official multi-subtask pipeline while replacing the original executor with EvoCUA-32B. Each chain contains 6 subtasks, and each subtask execution is limited to 10 steps. We run the pipeline on 7 configurations per domain, using 2 chains per configuration for \texttt{Calc} and 3 chains per configuration for \texttt{Impress}, \texttt{VLC}, and \texttt{VS Code}. This yields 84 raw level examples for \texttt{Calc} and 126 for the other three domains before final budget matching, after which we retain examples under the same per-domain budget used by our method.

\vspace{-0.05in}
\paragraph{ZeroGUI~\cite{yang2025zerogui}.}
ZeroGUI consists of two training stages: training with generated tasks and test-time training. For a fair comparison, we conduct only the first stage and exclude test-time training. Following the official implementation, we generate 10 instructions per round for 20 rounds per domain, yielding 200 candidate tasks per domain. We then randomly sample trajectories to match the same per-domain budget used by our method. We then follow its reward-based training recipe.

\vspace{-0.05in}
\paragraph{WebSTAR~\cite{webstar}.}
Since WebSTAR focuses on trajectory filtering given pre-collected trajectories, we apply it to our generated dataset using teacher trajectories from EvoCUA-32B. As our generation pipeline already includes a filtering stage, we exclude that component from our \modelname{}-GEN setting when applying WebSTAR. We follow the official WebSTAR implementation for the filtering procedure: each step is augmented with a generated thought, graded on a 0--10 scale, and retained only if the score exceeds 5. 

\section{Additional Experimental Results and Analysis}
\label{appendix:experiments}

\subsection{Statistics of Generated Data}
\label{appendix:statistics}
\begin{figure}[!b]
    \centering
    \begin{subfigure}[t]{0.495\textwidth}
        \centering
        \includegraphics[width=\linewidth]{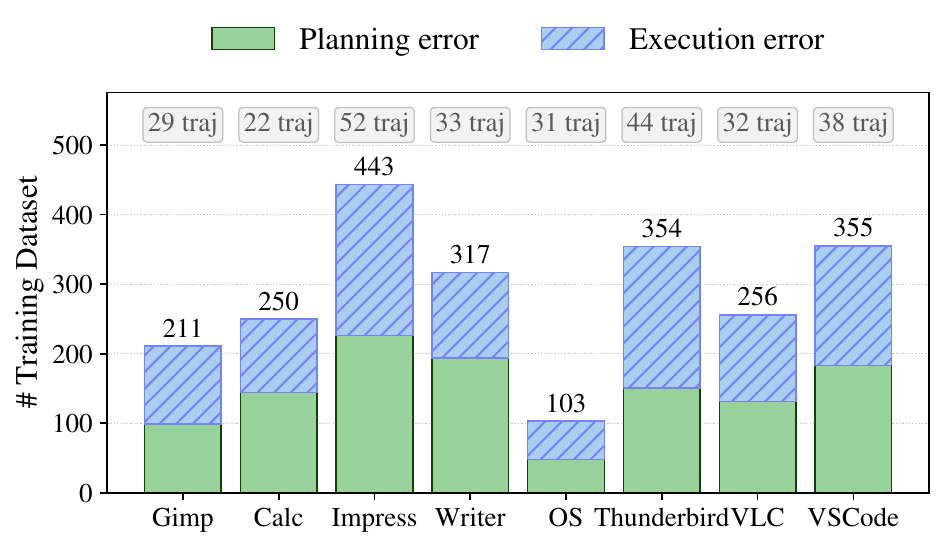}
        \vspace{-0.2in}
        \caption{EvoCUA-8B specialization.}
        \label{fig:stats_evocua8b_evocua32b}
    \end{subfigure}
    \hfill
    \begin{subfigure}[t]{0.495\textwidth}
        \centering
        \includegraphics[width=\linewidth]{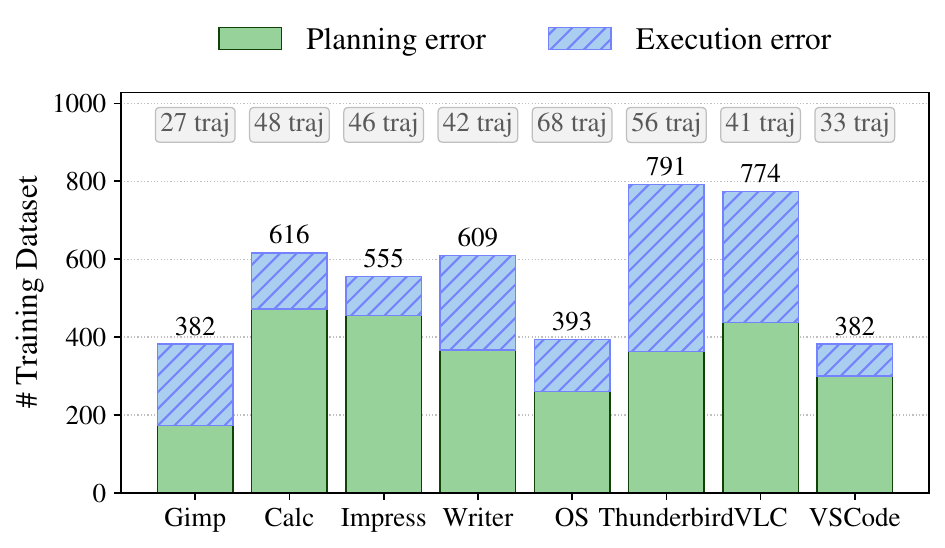}
        \vspace{-0.2in}
        \caption{OpenCUA-7B specialization.}
        \label{fig:stats_opencua7b_evocua32b}
    \end{subfigure}
    \caption{Domain-wise statistics of the generated specialization data for each model.}
    \label{fig:appendix_data_statistics}
\end{figure}

We report domain-level statistics of the generated datasets, including the number of teacher-pass and student-fail trajectories and the breakdown of planning and execution errors. These plots show that the generated supervision is highly heterogeneous across domains and student backbones. Some domains are dominated by planning-level discrepancies, whereas others contain a more balanced mixture of planning and execution errors. This heterogeneity is consistent with our specialization setting: different software domains expose different types of student failure, and the generated data reflects those domain-specific correction needs rather than a uniform error profile.

\subsection{Failure-Focused Trajectory Selection}

We compare three task-selection rules for building the specialization dataset: keeping all generated trajectories, retaining only tasks on which the teacher succeeds ($\pi_T$-pass), and retaining only tasks satisfying the same criterion used in \cref{subsec:method_datagen}, teacher-pass and student-fail ($\pi_T$-pass \& $\pi_S$-fail). This ablation tests whether our data-generation benefit comes simply from removing low-quality trajectories, or more specifically from concentrating supervision on unresolved student failures. As shown in \cref{tab:abl_filtering}, $\pi_T$-pass filtering alone is not sufficient and can even underperform using all generated trajectories on average. By contrast, $\pi_T$-pass \& $\pi_S$-fail filtering yields the strongest performance across all four domains. This result supports the core design of \modelname{}-GEN: the most useful specialization data is not generic successful behavior, but successful teacher behavior precisely where the current student still fails.

\begin{table}[h]
\centering
\caption{Comparison of task-selection rules for generated training data.}
\label{tab:abl_filtering}
\resizebox{0.75\linewidth}{!}{
\begin{tabular}{l cccc >{\columncolor{mygray}}c}
\toprule
~ & \textbf{Calc} & \textbf{Impress} & \textbf{VLC} & \textbf{VSCode} & \textbf{Avg.}\\
\midrule
Zero-shot & 28.07 & 37.66 & 45.71 & 51.30 & 40.69\\
\midrule
All trajectories  & 34.57 & 39.72 & 47.06 & \textbf{73.91} & 48.82 \\
Filtering ($\pi_T$-pass) &  24.82 & 42.55 & 43.14 & 72.46 & 45.74 \\
Filtering ($\pi_T$-pass \& $\pi_S$-fail)  & \textbf{41.13} & \textbf{50.35} & \textbf{56.86} & 72.46 & \textbf{55.20}\\

\bottomrule
\end{tabular}
}
\end{table}

\subsection{Adapting Specialization to Different Output Format (UI-TARS-1.5-7B)}

We additionally test our specialization framework on UI-TARS-1.5-7B~\cite{uitarsone}, and observe whether it can be adapted to a student whose output format differs from the one assumed in \cref{sec:preliminaries}. Our main method assumes the structured action format $a_t=(r_t, s_t, e_t)$ and applies error-aware masking accordingly in \cref{subsec:method_train}. However, UI-TARS-1.5-7B exposes only reasoning and tool execution, without a separate action-description component. We therefore use a modified masking rule that most closely matches our original design under this constraint. For planning-level errors ($\epsilon_{\text{PLAN}}$), we apply loss to both reasoning and tool-execution tokens; for execution-level errors ($\epsilon_{\text{EXEC}}$), we mask the reasoning tokens and apply loss only to the tool-execution tokens. Because reasoning tokens are directly optimized in the planning-error case, this variant is not equivalent to our main training rule and may be affected by teacher-student differences in thought style.

\begin{table}[h!]
\centering
\caption{Domain specialization results of UI-TARS-1.5-7B on OSWorld.}
\label{tab:uitars}
\resizebox{0.7\linewidth}{!}{
\begin{tabular}{l cccc >{\columncolor{mygray}}c}
\toprule
~ & \textbf{Calc} & \textbf{Impress} & \textbf{OS} & \textbf{VSCode} & \textbf{Avg.} \\
\midrule
\rowcolor{myyellow}
EvoCUA-32B & 51.06 & 52.98 & 75.00 & 65.22 & 61.07 \\ 
UI-TARS-1.5-7B & 7.09	& 21.98	& 16.67	& 30.43 & 19.04 \\
\rowcolor{myblue}
UI-TARS-1.5-7B \textbf{+ Ours} & 8.51 & 22.70 & 33.33 & 40.57 & 26.28 \\
\hspace{1em}$\Delta$ & \textbf{+1.42} & \textbf{+0.72} & \textbf{+16.66} & \textbf{+10.14} & \textbf{+7.24}\\
\bottomrule
\end{tabular}
}
\end{table}

As shown in \cref{tab:uitars}, this modified training rule still improves UI-TARS-1.5-7B on all four evaluated domains, with the largest gains on \texttt{OS} and \texttt{VSCode}. The improvements are smaller than those observed for EvoCUA and OpenCUA, which is reasonable because the original masking design is tailored to models with an explicit $r_t$--$s_t$--$e_t$ decomposition, whereas the UI-TARS variant must also supervise thought tokens in some cases. We therefore view this result as preliminary evidence that the framework can be adapted beyond our main output format, rather than as a direct like-for-like validation of the original training objective.

\clearpage
\section{Prompt Templates}
\label{appendix:prompt_templates}
\noindent
This section shows the main prompt templates used in \modelname{}-GEN. 
We use GPT-5-mini~\cite{openai2026gpt54miniNano} with following prompts during trajectory verification, weakness summarization, screenshot ranking, and query generation.

\begin{figure}[h]
\centering
\begin{center}
\begin{promptboxfootnotesize}[]
You are a strict verifier for one GUI task.
You will receive the task instruction and an agent step sequence.
For each step, data is ordered as: Thinking/Response -> Executed Action -> Screenshot.

## Task instruction
<instruction>

## Your job
1) Analyze the task instruction and set the criteria for task completion.
   All tasks in the instruction should be completed to get the pass.
2) Decide if the agent correctly completed the task objective (pass/fail).
3) If fail, provide a SHORT reason (3-4 sentences), concrete and behavior-focused.
   This should be detailed enough to help the agent improve without seeing the trajectory.
   Include which sub-task it failed, which component it did not ground correctly,
   or why the progress got stuck.

Return STRICT JSON only, with this exact schema:
{
  "task_completion_criteria": "list of task completion criteria",
  "verification_process": "check the task completion criteria one by one based on the trajectory and screenshots.",
  "agent_pass": true or false,
  "agent_failure_analysis": "detailed reason in 3-4 sentences; use empty string if agent_pass is true"
}

Rules:
- Pass only if the agent completed all required tasks in the instruction correctly.
- Be careful that the model may think the right plan in thought but the wrong action in execution.
- Do not trust self-reports like "done", "completed", or DONE action by themselves.
- Judge by actual trajectory behavior and screenshot evidence.
- Do not rely on literal PASS/FAIL labels in terminate messages.
- Be concise. English only in JSON values.
\end{promptboxfootnotesize}
\end{center}
\captionof{figure}{Trajectory verification prompt.}
\label{fig:appendix_prompt_verification}
\end{figure}

\begin{figure}[h]
\centering
\begin{center}
\begin{promptboxfootnotesize}[]
You are analyzing failure patterns of a student UI agent.
Input cases are tasks where TEACHER passed but STUDENT failed.

For each case, you receive:
- instruction
- student_agent_failure_analysis (judge explanation)

Please produce a concise, high-level report in JSON with this schema:
{
  "overall_summary": "string",
  "failure_categories": [
    {
      "category": "string",
      "what_student_cannot_do": "string",
      "likely_failed_features_or_operations": ["string"]
    }
  ]
}

Requirements:
1) Focus on sub-tasks the agent cannot do reliably.
2) Identify concrete operations the agent misuses or fails to execute.
3) Categories should be notably different from each other.
4) Group repeated failures into reusable categories.
5) Do not include markdown; return JSON only.
\end{promptboxfootnotesize}
\end{center}
\captionof{figure}{Teacher--student weakness summarization prompt.}
\label{fig:appendix_prompt_weakness}
\end{figure}

\begin{figure}[h]
\centering
\begin{center}
\begin{promptboxfootnotesize}[]
You are evaluating screenshots from a single software domain.
Goal: select the screenshots that maximize understanding of the domain's
features and UI components.

You will receive candidate screenshots in this pattern:
Image 0: <image>
Image 1: <image>
...
Use the number in each "Image N" label as the index.

Select exactly K screenshots.
Prioritize:
1) Coverage of distinct major features/workflows.
2) Diversity of visible UI components/layout states.
3) Informational richness (settings/panels/dialogs/menus/output views).
Avoid near-duplicates and low-information transitional frames.

Return ONLY valid JSON with this schema:
{
  "selected_indices": [int, ...],
  "reasons": [
    {
      "index": int,
      "reason": "short reason focused on coverage value"
    }
  ]
}
\end{promptboxfootnotesize}
\end{center}
\captionof{figure}{Screenshot ranking prompt.}
\label{fig:appendix_prompt_screenshot}
\end{figure}

\begin{figure}[h]
\centering
\begin{center}
\begin{promptboxfootnotesize}[]
Goal:
- Propose new task instructions that specifically improve abilities where the student still fails.
- Keep tasks realistic for the given config environment.
- Obey the Workspace / path contract; do not name repos, folders, or files
  that this docker setup does not open or download.
- While instructions are grounded in the config and failure analysis,
  they should be diverse and distinct from the prior instructions.
- Do not duplicate or lightly paraphrase the prior instructions.
- Tasks should not be tutorials or step-by-step guides.
- Tasks should be easy, concise, and possible to finish in a few steps.

Input context:
1) Prior instructions already used (avoid overlap/paraphrase)
2) Student weakness analysis (teacher pass, student fail)
3) Workspace / path contract
4) Current docker config array to target
5) Extra file/folder/code context from this config (provide_info)

Requirements:
- Generate exactly Y instructions.
- Each instruction must be concise end-user style English.
- Do not include more than two simple and easy sub-tasks.
- Every instruction must satisfy the workspace/path contract.
- Must target one or more weak abilities from the analysis.
- Must be feasible with this config and attached context.
- Instructions should be less than 15 words long.
- Generated tasks should be significantly different from each other.

Return STRICT JSON only:
{
  "queries": [
    {
      "reference_config_id": "string",
      "instruction": "string",
      "targets_student_gaps": ["string"],
      "rationale": "one short sentence"
    }
  ]
}
\end{promptboxfootnotesize}
\end{center}
\captionof{figure}{Query-generation prompt \texttt{with weakness report}.}
\label{fig:appendix_prompt_generation_weak}
\end{figure}

\begin{figure}[p]
\centering
\begin{center}
\begin{promptboxfootnotesize}[]
Goal:
- Propose new task instructions by exploring what appears in screenshots.
- Prioritize new features/workflows/subtasks not present in prior instructions.
- Keep tasks easy, short, and realistic for the given config environment.
- Section 3 (Workspace / path contract) overrides screenshots.
- While instructions are grounded in the config and the screenshots,
  they should be diverse and distinct from the prior instructions.
- Do not duplicate or lightly paraphrase the prior instructions.
- Tasks should not be tutorials or step-by-step guides.
- Tasks should be easy, concise, and possible to finish in a few steps.

Input context:
1) Prior instructions already used (avoid overlap/paraphrase)
2) Student weakness analysis: (Not used in this run.)
3) Workspace / path contract
4) Current docker config array to target
5) Extra file/folder/code context from this config (provide_info)

Requirements:
- Generate exactly Y instructions.
- Each instruction must be concise end-user style English.
- Do not include more than two simple and easy sub-tasks.
- Every instruction must satisfy the workspace/path contract.
- Must maximize diversity and novelty versus prior instructions.
- Each query object must include only reference_config_id, instruction, and rationale.
- Instructions should be less than 15 words long.
- Generated tasks should be significantly different from each other.

Return STRICT JSON only:
{
  "queries": [
    {
      "reference_config_id": "string",
      "instruction": "a short string",
      "rationale": "one short sentence"
    }
  ]
}
\end{promptboxfootnotesize}
\end{center}
\captionof{figure}{Query-generation prompt \texttt{without weakness report}.}
\label{fig:appendix_prompt_generation_open}
\end{figure}

\clearpage
\clearpage
\section{Qualitative Results}
\label{appendix:qualitative}

\newcommand{\tracepromptopts}{listing options={basicstyle=\ttfamily\tiny,breaklines=true,columns=fullflexible}}

\subsection{Weakness Report and Synthetic Query Results}
\noindent
This section shows how the weakness-reporting stage is connected to the
subsequent synthetic-query generation stage from the \texttt{libreoffice\_calc} domain. 
For each example, we first present a formatted excerpt from the weakness report, then show representative synthetic queries derived from the identified failure categories, followed by a brief analysis of the report-to-query linkage.

\begin{figure}[h]
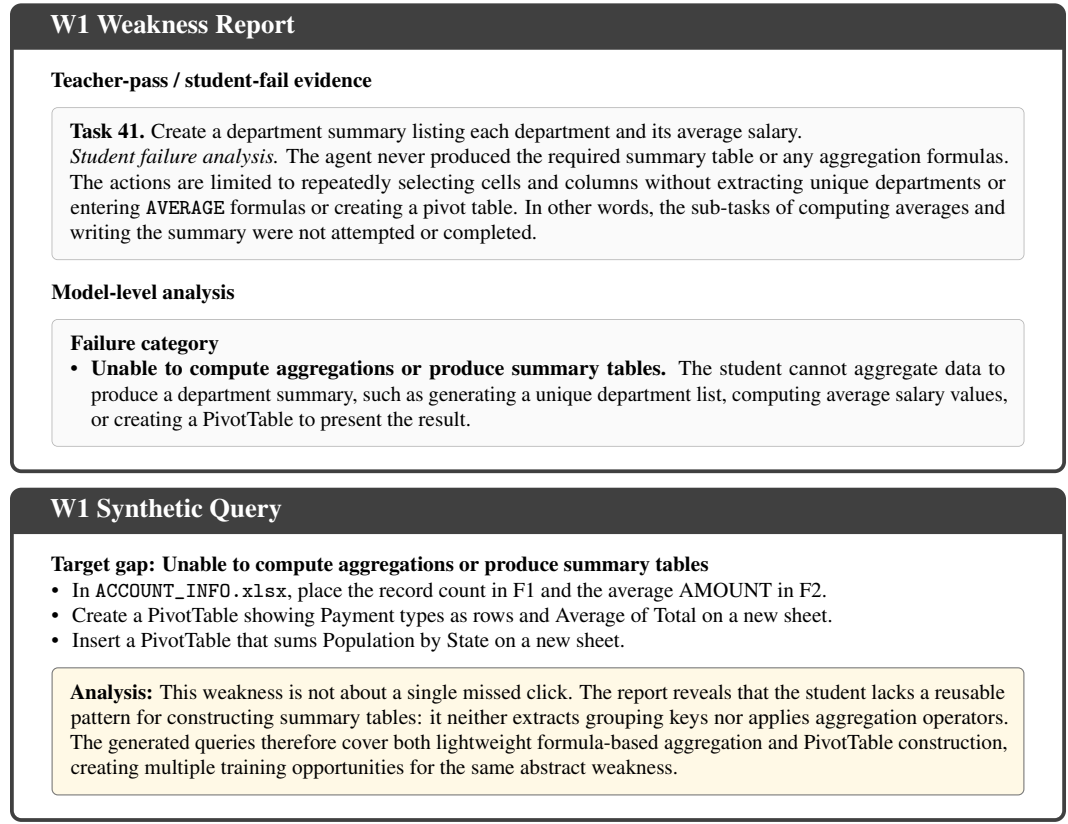

\begin{tcolorbox}[
  title={\textbf{W1 Weakness Report}},
  colback=white,
  fontupper=\footnotesize,
  width=\textwidth
]
\textbf{Teacher-pass / student-fail evidence}
\begin{tcolorbox}[colback=black!2, colframe=black!30, boxrule=0.3pt, arc=2pt, left=4pt, right=4pt, top=3pt, bottom=3pt]
\textbf{Task 41.} Create a department summary listing each department and its average salary.\\
\textit{Student failure analysis.} The agent never produced the required summary table or any aggregation formulas. The actions are limited to repeatedly selecting cells and columns without extracting unique departments or entering \texttt{AVERAGE} formulas or creating a pivot table. In other words, the sub-tasks of computing averages and writing the summary were not attempted or completed.
\end{tcolorbox}

\vspace{0.4em}
\textbf{Model-level analysis}
\begin{tcolorbox}[colback=black!2, colframe=black!30, boxrule=0.3pt, arc=2pt, left=4pt, right=4pt, top=3pt, bottom=3pt]
\textbf{Failure category}
\begin{itemize}[leftmargin=*, nosep]
    \item \textbf{Unable to compute aggregations or produce summary tables.} The student cannot aggregate data to produce a department summary, such as generating a unique department list, computing average salary values, or creating a PivotTable to present the result.
\end{itemize}
\end{tcolorbox}
\end{tcolorbox}

\begin{tcolorbox}[
  title={\textbf{W1 Synthetic Query}},
  colback=white,
  fontupper=\footnotesize,
  width=\textwidth
]
\textbf{Target gap: Unable to compute aggregations or produce summary tables}
\begin{itemize}[leftmargin=*, nosep]
\item In \texttt{ACCOUNT\_INFO.xlsx}, place the record count in F1 and the average AMOUNT in F2.
\item Create a PivotTable showing Payment types as rows and Average of Total on a new sheet.
\item Insert a PivotTable that sums Population by State on a new sheet.
\end{itemize}

\begin{tcolorbox}[
    colback=myyellow,
    colframe=black!55,
    boxrule=0.4pt,
    arc=2pt,
    left=4pt,
    right=4pt,
    top=3pt,
    bottom=3pt
]
\textbf{Analysis:}
This weakness is not about a single missed click. The report reveals that the
student lacks a reusable pattern for constructing summary tables: it neither
extracts grouping keys nor applies aggregation operators. The generated
queries therefore cover both lightweight formula-based aggregation and
PivotTable construction, creating multiple training opportunities for the same
abstract weakness.
\end{tcolorbox}
\end{tcolorbox}
\caption{Weakness Report and Synthetic Queries: Example \#1}
\end{figure}

\clearpage
\begin{figure}[H]
\begin{tcolorbox}[
  title={\textbf{W2 Weakness Report}},
  colback=white,
  fontupper=\footnotesize,
  width=\textwidth
]
\textbf{Teacher-pass / student-fail evidence}
\begin{tcolorbox}[colback=black!2, colframe=black!30, boxrule=0.3pt, arc=2pt, left=4pt, right=4pt, top=3pt, bottom=3pt]
\textbf{Task 12.} Freeze the header row and set the used range as the print area.\\
\textit{Student failure analysis.} The agent did not complete the print-area subtask: after selecting the used range there is no evidence it set the range as the print area. The agent also applied a freeze, but the freeze line is below row 2 rather than immediately below the header row, so the header-only freeze was not precisely applied.
\end{tcolorbox}

\begin{tcolorbox}[colback=black!2, colframe=black!30, boxrule=0.3pt, arc=2pt, left=4pt, right=4pt, top=3pt, bottom=3pt]
\textbf{Task 16.} Freeze the header row and first column in \texttt{Order\_Details.xlsx}.\\
\textit{Student failure analysis.} The agent failed to select the required freeze origin (\texttt{B2}) before applying the Freeze command. The Name box in the screenshots shows \texttt{A1} throughout, so the wrong cell remained active and the required row-and-column freeze could not be applied reliably.
\end{tcolorbox}

\vspace{0.4em}
\textbf{Model-level analysis}
\begin{tcolorbox}[colback=black!2, colframe=black!30, boxrule=0.3pt, arc=2pt, left=4pt, right=4pt, top=3pt, bottom=3pt]
\textbf{Overall summary.} Across these cases, the student UI agent reliably navigates and clicks cells but fails at higher-level spreadsheet operations that require setting the correct active selection and invoking specific commands or dialogs.

\vspace{0.3em}
\textbf{Failure categories}
\begin{itemize}[leftmargin=*, nosep]
    \item \textbf{Failure to apply print-area settings after range selection.} The student can select a used range but fails to execute the Print Area command to define that range as the sheet's print area.
    \item \textbf{Incorrect freeze/pane origin selection.} The student cannot freeze exactly the intended rows and columns because it does not activate the correct origin cell before applying Freeze operations.
\end{itemize}
\end{tcolorbox}
\end{tcolorbox}

\begin{tcolorbox}[
  title={\textbf{W2 Synthetic Query}},
  colback=white,
  fontupper=\footnotesize,
  width=\textwidth
]
\textbf{Target gap: Failure to apply print-area settings after range selection}
\begin{itemize}[leftmargin=*, nosep]
    \item Set print area to A1:D12 and confirm via Print Preview.
    \item Set columns A:D as the print area and open Print Preview.
\end{itemize}

\vspace{0.3em}
\textbf{Target gap: Incorrect freeze/pane origin selection}
\begin{itemize}[leftmargin=*, nosep]
    \item Freeze panes so rows 1--2 and columns A--B remain visible when scrolling.
    \item Select cell C2 then freeze panes so top rows and left columns remain visible.
\end{itemize}

\vspace{0.3em}
\textbf{Target gap: Unable to compute aggregations or produce summary tables}
\begin{itemize}[leftmargin=*, nosep]
    \item In \texttt{ACCOUNT\_INFO.xlsx}, place the record count in F1 and the average AMOUNT in F2.
    \item Create a PivotTable showing Payment types as rows and Average of Total on a new sheet.
\end{itemize}

\vspace{0.5em}
\begin{tcolorbox}[
    colback=myyellow,
    colframe=black!55,
    boxrule=0.4pt,
    arc=2pt,
    left=4pt,
    right=4pt,
    top=3pt,
    bottom=3pt
]
\textbf{Analysis:}
The weakness report identifies two operational failures: the student does not
commit the selected range as a print area, and it applies freeze panes from
the wrong active cell. The generated synthetic queries then isolate those
exact operations into short practice tasks, repeatedly requiring explicit
print-area definition or a specific freeze origin such as \texttt{B2} or
\texttt{C2}.
\end{tcolorbox}
\end{tcolorbox}
\caption{Weakness Report and Synthetic Queries: Example \#2}
\end{figure}

\begin{figure}[H]
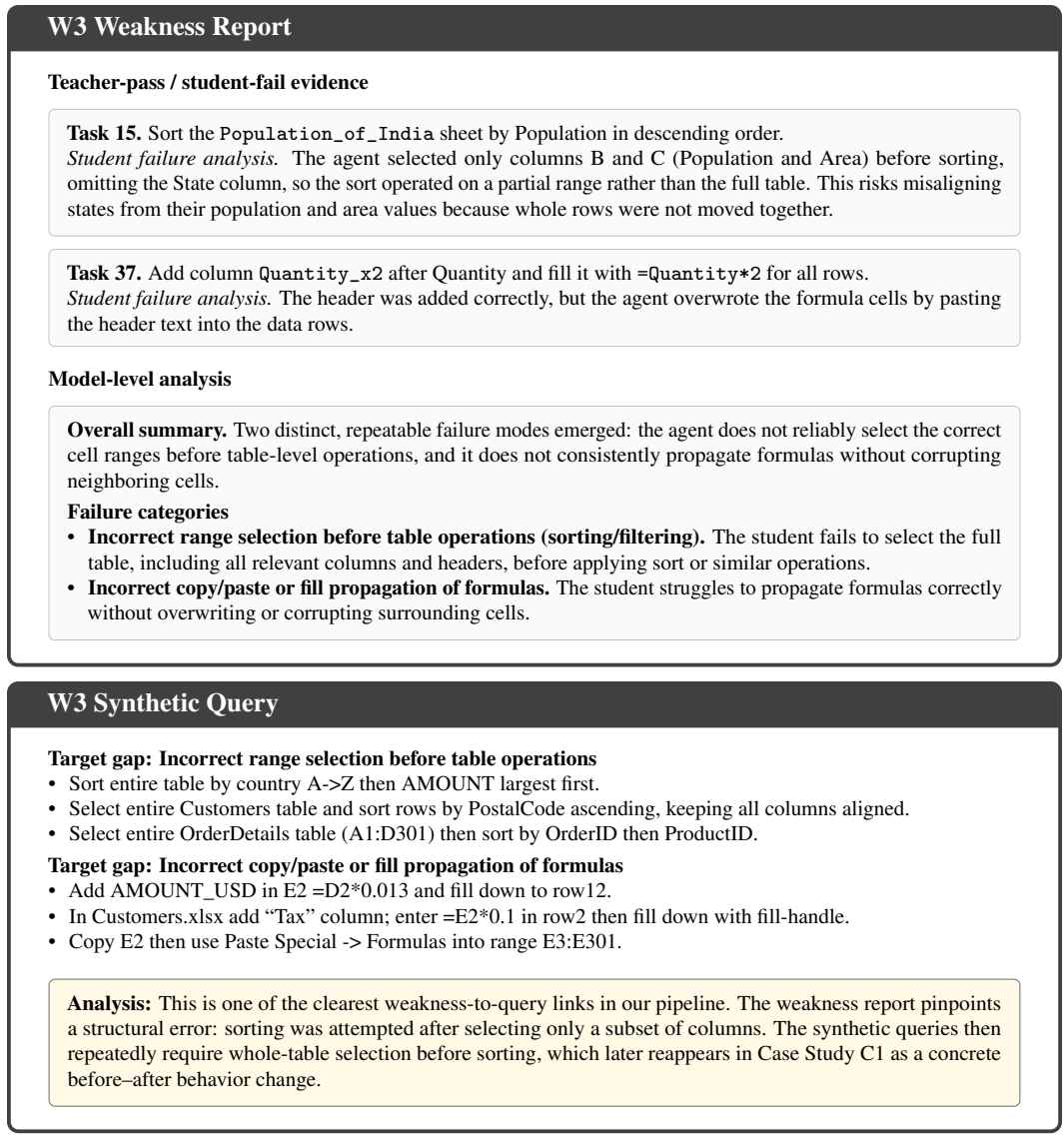

\begin{tcolorbox}[
  title={\textbf{W3 Weakness Report}},
  colback=white,
  fontupper=\footnotesize,
  width=\textwidth
]
\textbf{Teacher-pass / student-fail evidence}
\begin{tcolorbox}[colback=black!2, colframe=black!30, boxrule=0.3pt, arc=2pt, left=4pt, right=4pt, top=3pt, bottom=3pt]
\textbf{Task 15.} Sort the \texttt{Population\_of\_India} sheet by Population in descending order.\\
\textit{Student failure analysis.} The agent selected only columns B and C (Population and Area) before sorting, omitting the State column, so the sort operated on a partial range rather than the full table. This risks misaligning states from their population and area values because whole rows were not moved together.
\end{tcolorbox}

\begin{tcolorbox}[colback=black!2, colframe=black!30, boxrule=0.3pt, arc=2pt, left=4pt, right=4pt, top=3pt, bottom=3pt]
\textbf{Task 37.} Add column \texttt{Quantity\_x2} after Quantity and fill it with \texttt{=Quantity*2} for all rows.\\
\textit{Student failure analysis.} The header was added correctly, but the agent overwrote the formula cells by pasting the header text into the data rows.
\end{tcolorbox}

\vspace{0.4em}
\textbf{Model-level analysis}
\begin{tcolorbox}[colback=black!2, colframe=black!30, boxrule=0.3pt, arc=2pt, left=4pt, right=4pt, top=3pt, bottom=3pt]
\textbf{Overall summary.} Two distinct, repeatable failure modes emerged: the agent does not reliably select the correct cell ranges before table-level operations, and it does not consistently propagate formulas without corrupting neighboring cells.

\vspace{0.3em}
\textbf{Failure categories}
\begin{itemize}[leftmargin=*, nosep]
    \item \textbf{Incorrect range selection before table operations (sorting/filtering).} The student fails to select the full table, including all relevant columns and headers, before applying sort or similar operations.
    \item \textbf{Incorrect copy/paste or fill propagation of formulas.} The student struggles to propagate formulas correctly without overwriting or corrupting surrounding cells.
\end{itemize}
\end{tcolorbox}
\end{tcolorbox}

\begin{tcolorbox}[
    title={\textbf{W3 Synthetic Query}},
    colback=white,
    fontupper=\footnotesize,
    width=\textwidth
]

\textbf{Target gap: Incorrect range selection before table operations}
\begin{itemize}[leftmargin=*, nosep]
    \item Sort entire table by country A->Z then AMOUNT largest first.
    \item Select entire Customers table and sort rows by PostalCode ascending, keeping all columns aligned.
    \item Select entire OrderDetails table (A1:D301) then sort by OrderID then ProductID.
\end{itemize}

\vspace{0.3em}
\textbf{Target gap: Incorrect copy/paste or fill propagation of formulas}
\begin{itemize}[leftmargin=*, nosep]
    \item Add AMOUNT\_USD in E2 =D2*0.013 and fill down to row12.
    \item In Customers.xlsx add ``Tax'' column; enter =E2*0.1 in row2 then fill down with fill-handle.
    \item Copy E2 then use Paste Special -> Formulas into range E3:E301.
\end{itemize}

\vspace{0.5em}
\begin{tcolorbox}[
    colback=myyellow,
    colframe=black!55,
    boxrule=0.4pt,
    arc=2pt,
    left=4pt,
    right=4pt,
    top=3pt,
    bottom=3pt
]
\textbf{Analysis:}
This is one of the clearest weakness-to-query links in our pipeline. The
weakness report pinpoints a structural error: sorting was attempted after
selecting only a subset of columns. The synthetic queries then repeatedly
require whole-table selection before sorting, which later reappears in Case
Study C1 as a concrete before--after behavior change.
\end{tcolorbox}
\end{tcolorbox}
\caption{Weakness Report and Synthetic Queries: Example \#3}
\end{figure}

\newpage
\subsection{Case Study}
\newcolumntype{Y}{>{\centering\arraybackslash}X}
\noindent
The case studies compare OSWorld benchmark trajectories before and after specialization in \texttt{LibreOffice Calc} and \texttt{LibreOffice Impress} domains. Each step includes the agent's observation, shown as the screenshot visible at that step, together with the corresponding model response and a concise action summary. We omit the reasoning portion of the model response for brevity. These examples illustrate how specialization alters the model's local decision-making behavior, rather than only improving the final task outcome.

\begin{figure}[H]
\begin{tcolorbox}[
  colback=white,
  fontupper=\footnotesize,
  width=\textwidth,
  breakable
]
\textbf{Instruction}:
Could you help me to sort the records accoring to the amounts ascendingly?

\vspace{0.5em}
\begin{tcolorbox}[colback=red!4, colframe=black!55, boxrule=0.4pt, arc=2pt, left=4pt, right=4pt, top=3pt, bottom=3pt, breakable]
\textbf{Before Specialization} \\
\vspace{0.3em}
\begin{tabularx}{\linewidth}{@{}YYY@{}}
    Step 4 & \textbf{Step 5} & Step 6 \\

    \includegraphics[width=\linewidth]{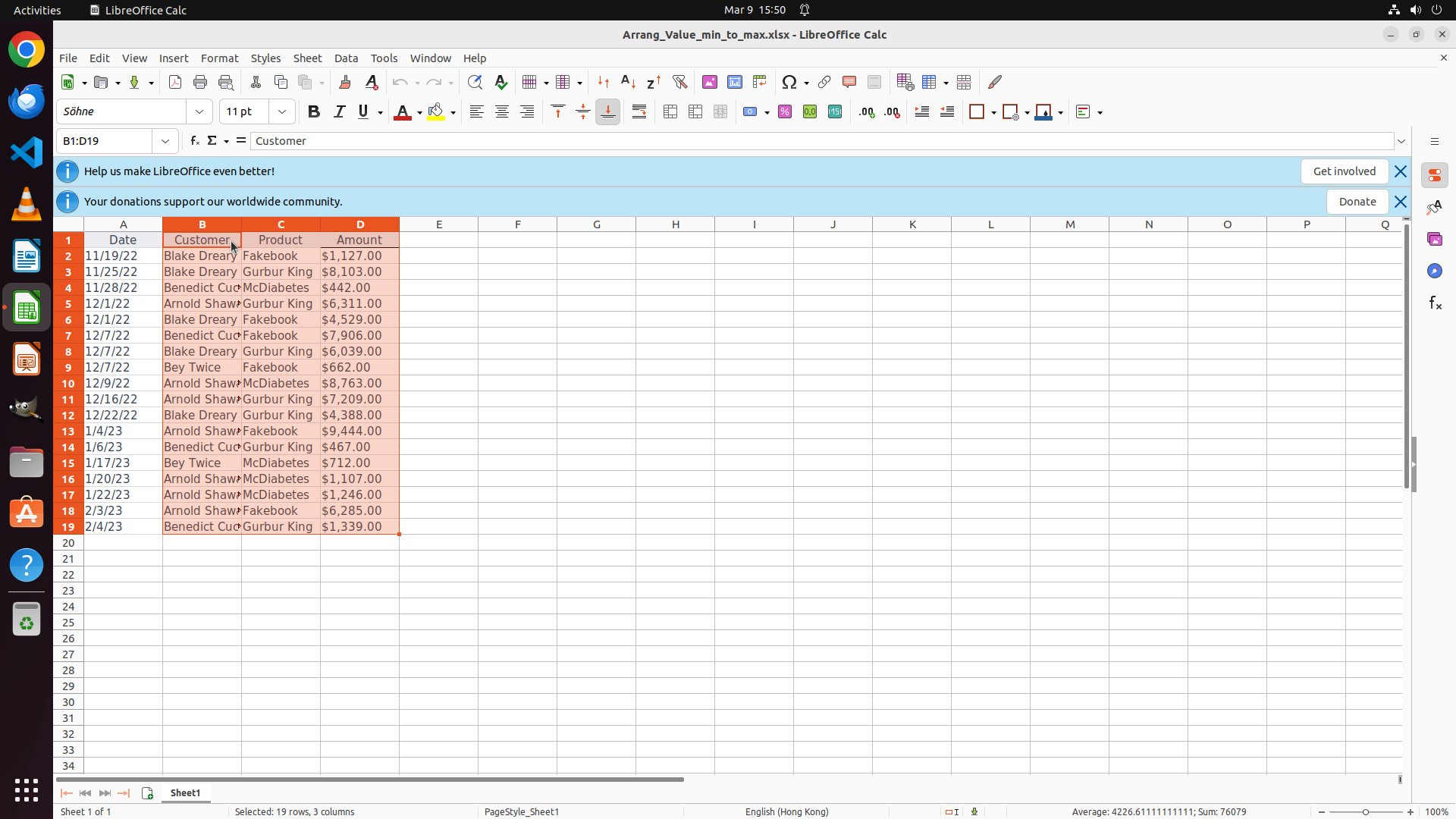}
    & \includegraphics[width=\linewidth]{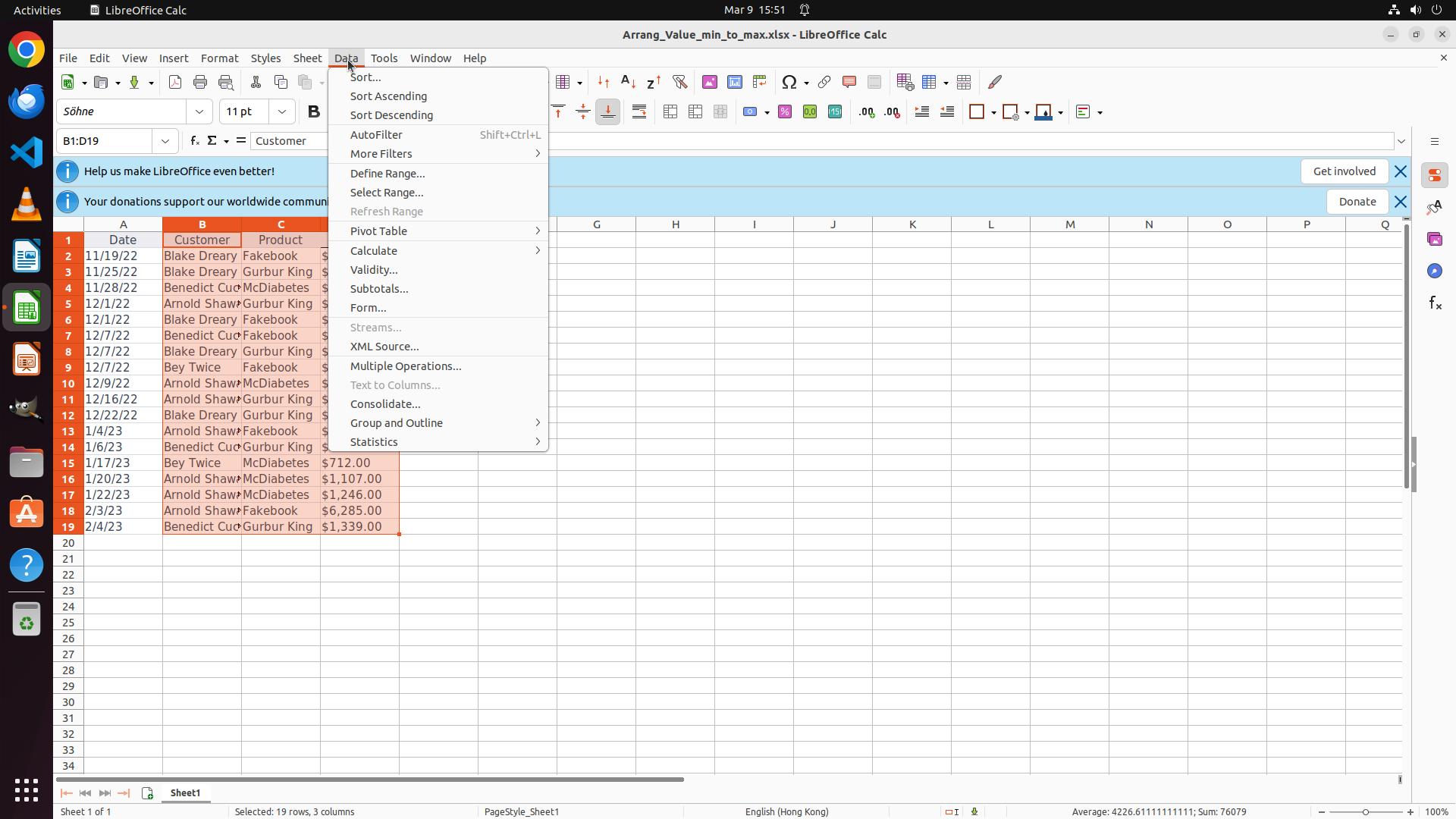}
    & \includegraphics[width=\linewidth]{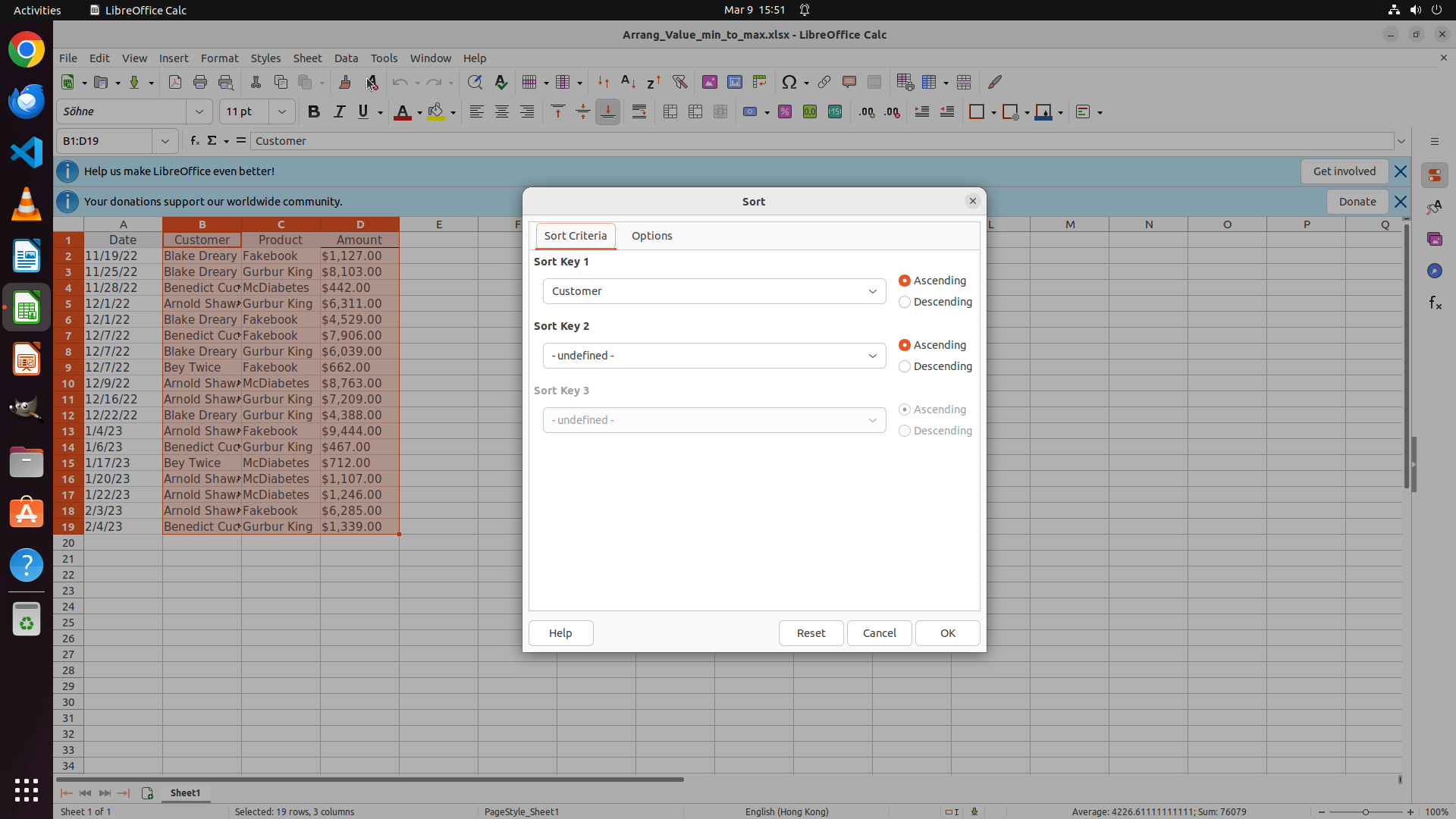} \\

    \scriptsize
    Clicked on cell B1 (Customer column) and selected the range from the
    current cell to the last populated cell in column~D.
    \par\texttt{HotKey(ctrl, shift, end)}
    &
    \scriptsize
    The data range B1:D19 is now selected, with ``Selected: 19 rows,
    3 columns'' shown in the status bar.
    \par\texttt{Click(455,75)}

    &
    \scriptsize
    Opened the Data menu and proceeded to the Sort dialog.
    \par\texttt{Click(480,99)}
\end{tabularx}

\textit{The baseline model selects only columns B--D, so the sort is applied to a partial table rather than the full sheet range.}
\end{tcolorbox}

\begin{tcolorbox}[colback=mygreen!8, colframe=black!55, boxrule=0.4pt, arc=2pt, left=4pt, right=4pt, top=3pt, bottom=3pt, breakable]
\textbf{After Specialization}

\vspace{0.3em}
\begin{tabularx}{\linewidth}{@{}YYY@{}}
    Step 4 & \textbf{Step 5} & Step 6 \\

    \includegraphics[width=\linewidth]{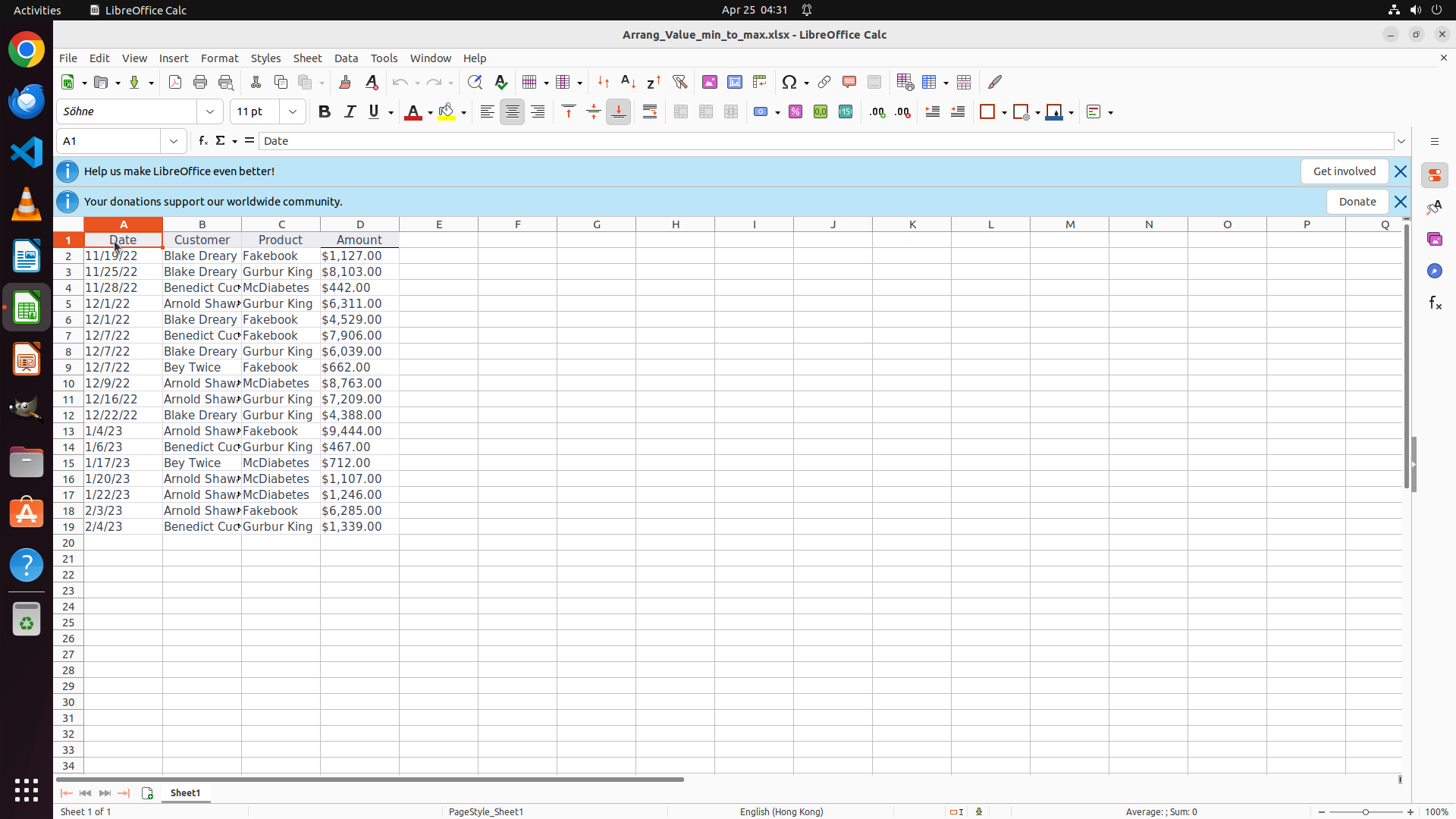}
    & \includegraphics[width=\linewidth]{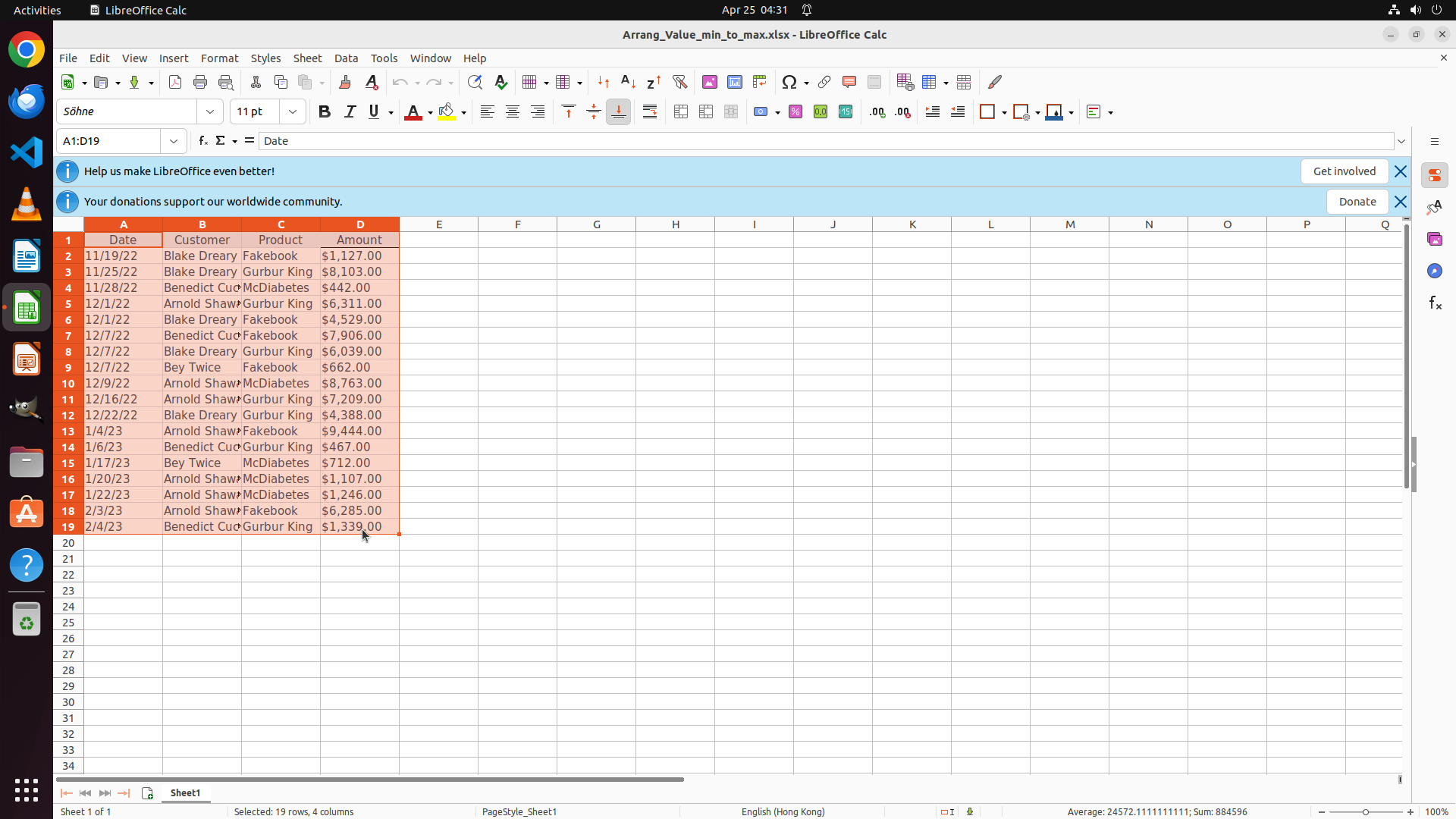}
    & \includegraphics[width=\linewidth]{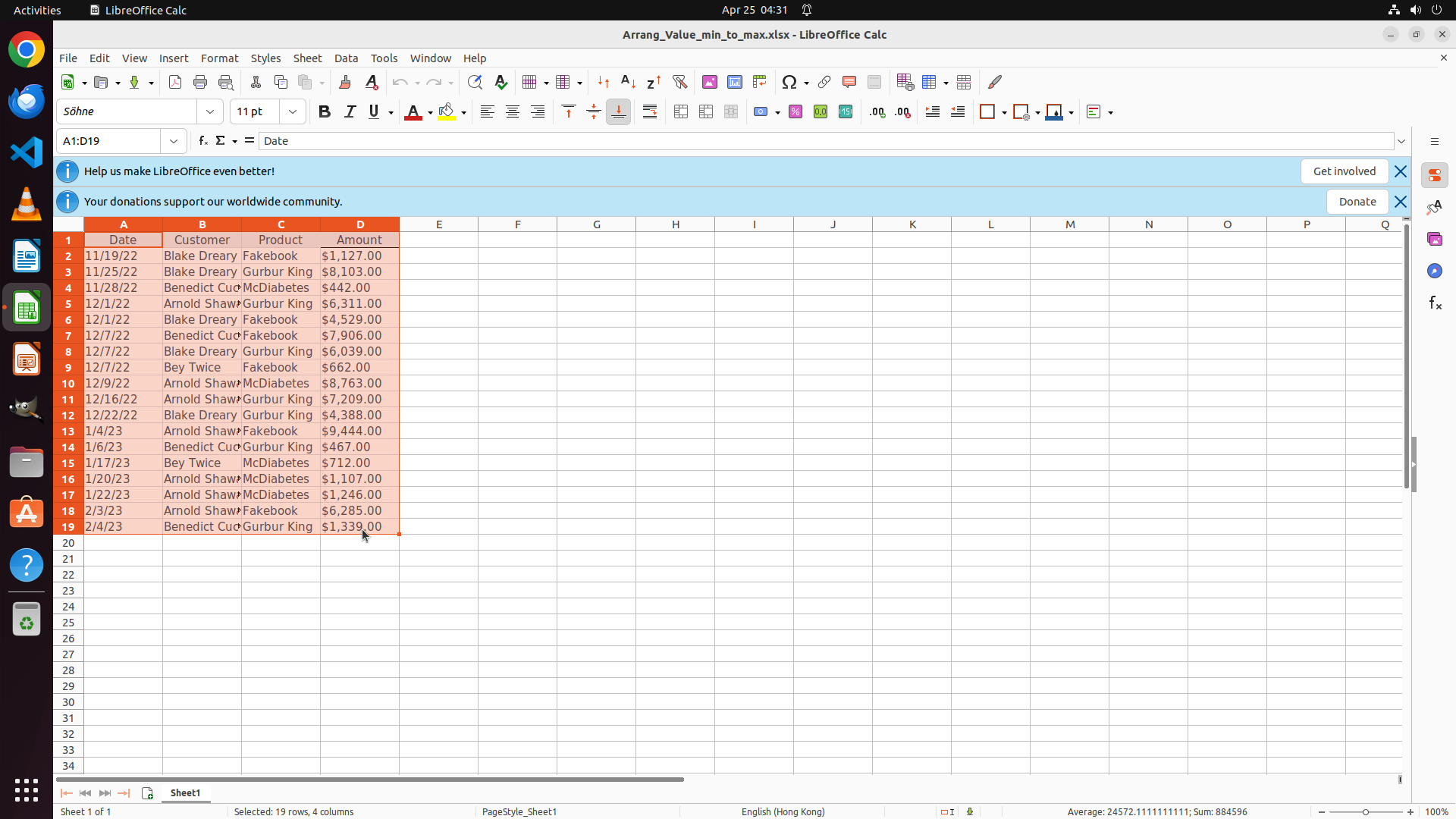} \\

    \scriptsize
    Clicked on cell A1 and selected the entire data range, including all columns
    (Date, Customer, Product, Amount) and all rows with data.
    \par\texttt{KeyDown(Shift)}, \texttt{Click(474,694)}, \texttt{KeyUp(Shift)}

    &
    \scriptsize
    The data range A1:D19 is now selected, with ``Selected: 19 rows,
    4 columns'' shown in the status bar.
    \par\texttt{Click(455,75)}

    &
    \scriptsize
    Opened the Data menu and proceeded to the Sort dialog.
    \par\texttt{Click(480,99)}
\end{tabularx}
\textit{The specialized model explicitly expands the selection to A1:D19 before proceeding to sorting.}
\end{tcolorbox}

\begin{tcolorbox}[
    colback=myyellow,
    colframe=black!55,
    boxrule=0.4pt,
    arc=2pt,
    left=4pt,
    right=4pt,
    top=3pt,
    bottom=3pt
]
\textbf{Analysis:}
The decisive difference is the selected range. The failing baseline trajectory
sorts only \texttt{B1:D19}, excluding the leftmost column and therefore
violating the full-table sorting requirement. The specialized model instead
selects \texttt{A1:D19}, preserving row integrity during sorting. This
behavioral change matches the step-4 weakness category shown above.
\end{tcolorbox}
\end{tcolorbox}
\caption{Case Study \#1 (Domain: \texttt{Libreoffice Calc})}
\end{figure}

\newpage

\begin{figure}[H]
\begin{tcolorbox}[
  colback=white,
  fontupper=\footnotesize,
  width=\textwidth,
  breakable
]
\textbf{Instruction}: 
Help me fill in the Gross profit column by subtracting all the available
expenses including discounts, allowances, material and labor charges, and
overhead from the actual sale, i.e., the sales after deducting the returns.
Then under column A named "Year\_Profit" in a new sheet "Sheet2", display the
Year Column in Sheet 1 as text appended by a "\_" with the corresponding
integer digits of Gross Profit value.

\vspace{0.5em}
\begin{tcolorbox}[colback=red!4, colframe=black!55, boxrule=0.4pt, arc=2pt, left=4pt, right=4pt, top=3pt, bottom=3pt, breakable]
\textbf{Before Specialization} \\
\vspace{0.3em}
\begin{tabularx}{\linewidth}{@{}YYY@{}}
    Step 47 & Step 48 & \textbf{Step 49} \\

    \includegraphics[width=\linewidth]{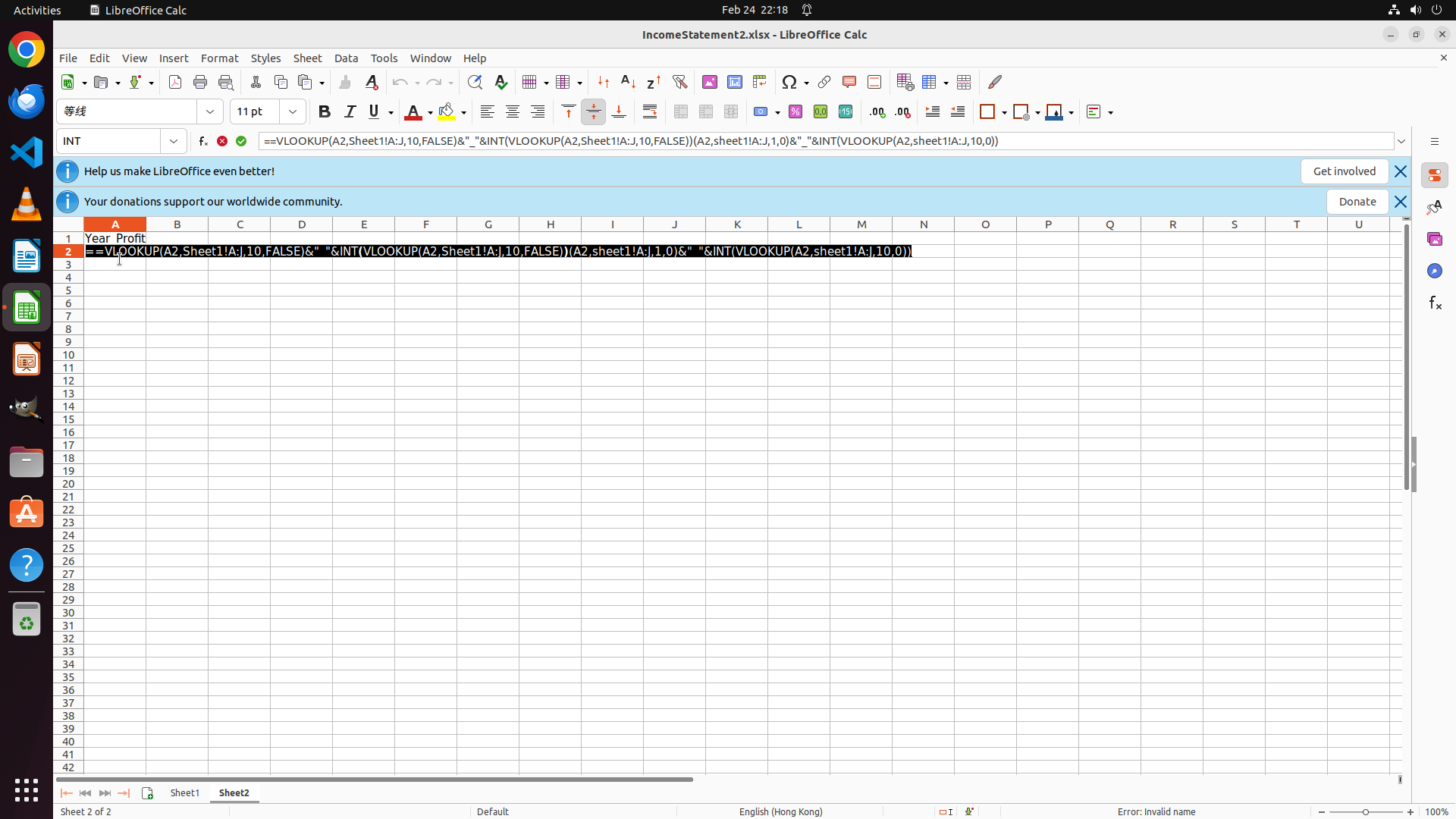}
    & \includegraphics[width=\linewidth]{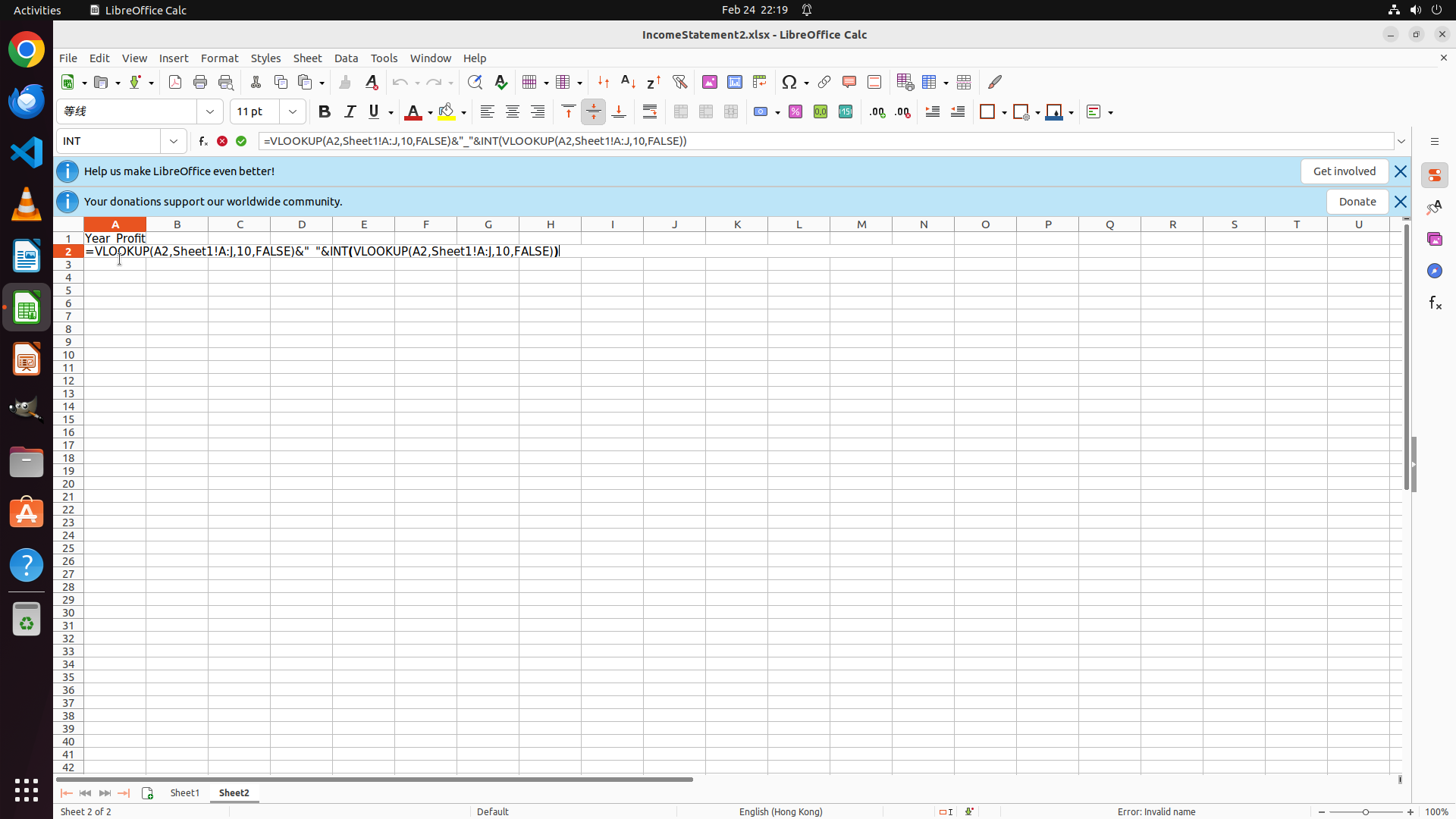}
    & \includegraphics[width=\linewidth]{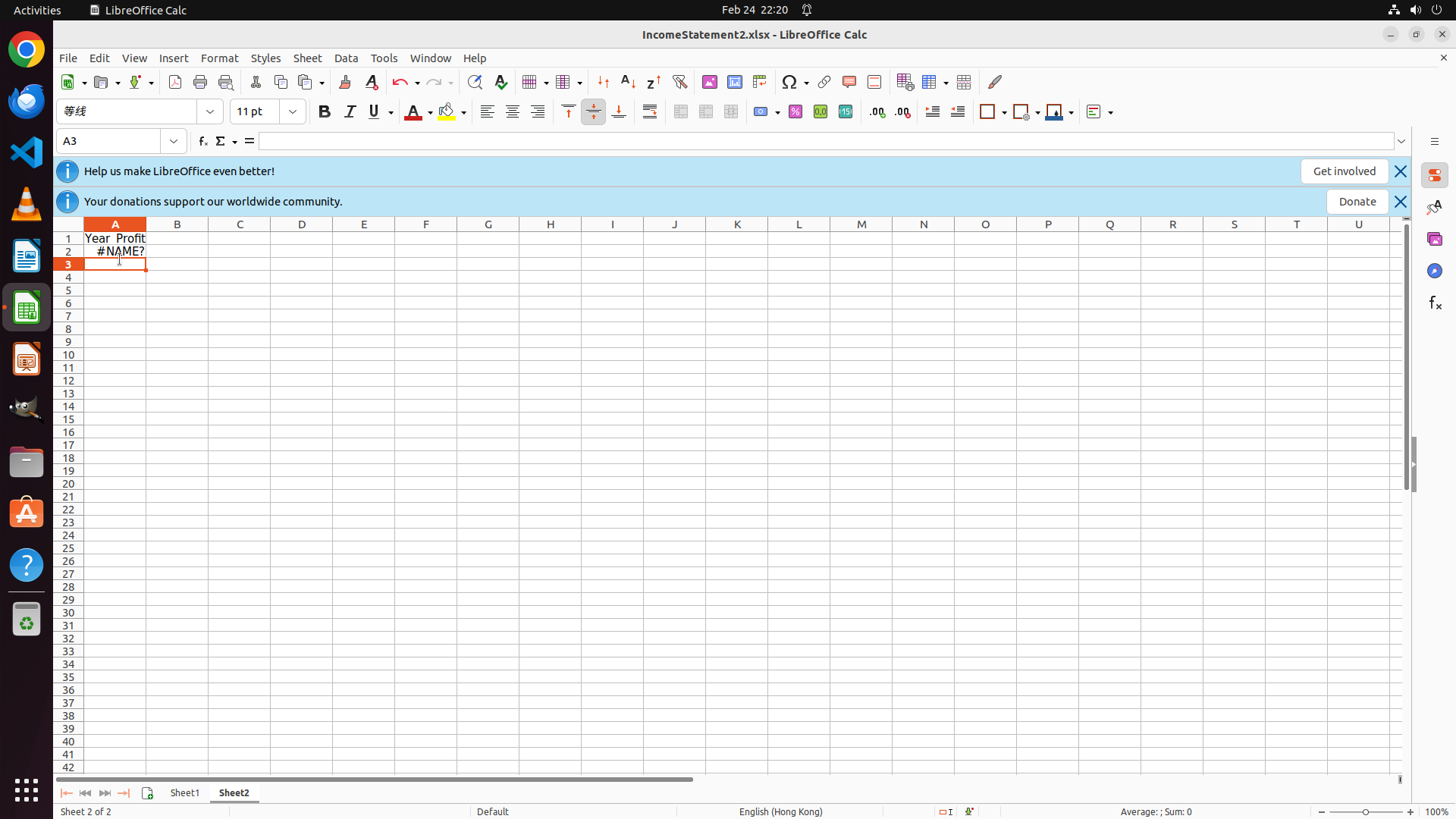} \\

    \scriptsize
    Selected the formula bar contents and prepared to replace them with a
    lookup-style Year\_Profit formula.
    \par\texttt{HotKey(ctrl,a)}
    &
    \scriptsize
    Typed a long \texttt{VLOOKUP}-based concatenation formula into cell~A2.
    \par\texttt{Type(=VLOOKUP(...)\&"\_"\&)}
    \par\texttt{INT(VLOOKUP(...)))}
    &
    \scriptsize
    Executed the formula, which resulted in a \texttt{\#NAME?} error in
    \texttt{Sheet2!A2}.
    \par\texttt{Key(enter)}
\end{tabularx}

\textit{The baseline commits to an unnecessarily complex lookup-based construction and ends with a \texttt{\#NAME?} error instead of a usable row-wise formula.}

\end{tcolorbox}
\vspace{0.5em}
\begin{tcolorbox}[colback=mygreen!8, colframe=black!55, boxrule=0.4pt, arc=2pt, left=4pt, right=4pt, top=3pt, bottom=3pt, breakable]
\textbf{After Specialization}

\vspace{0.3em}
\begin{tabularx}{\linewidth}{@{}YYY@{}}
    Step 13 & Step 14 & \textbf{Step 15} \\

    \includegraphics[width=\linewidth]{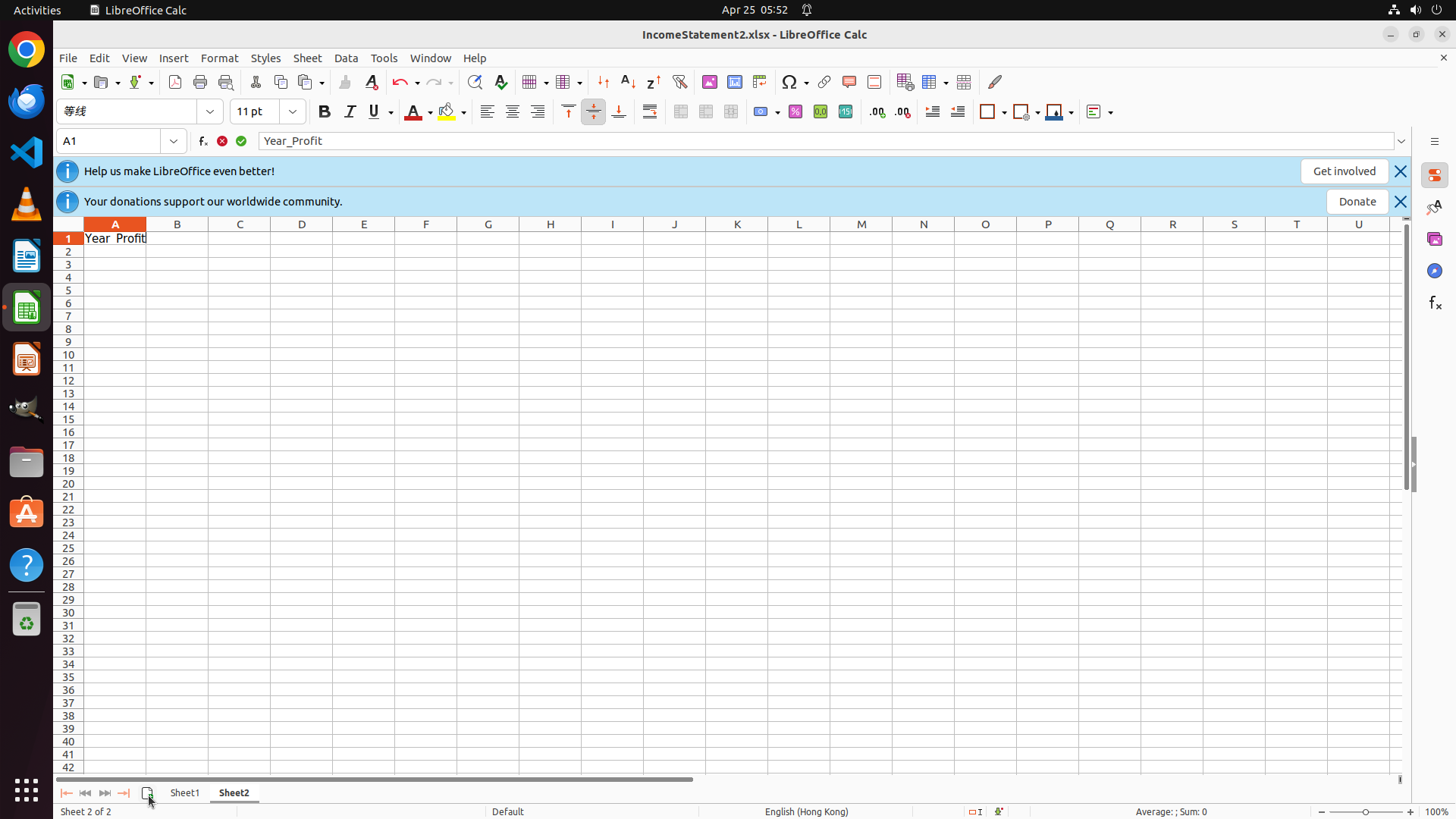}
    & \includegraphics[width=\linewidth]{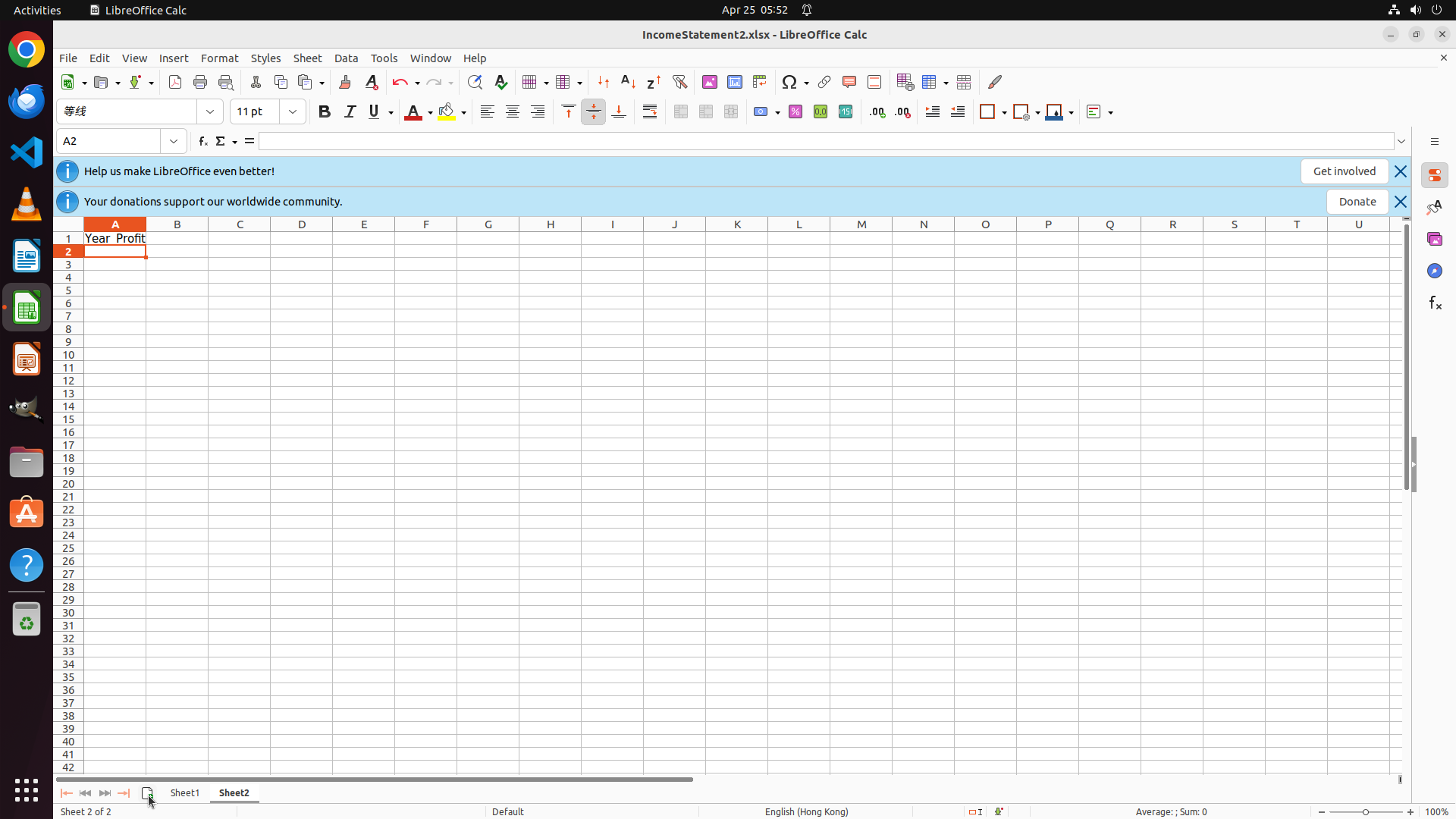}
    & \includegraphics[width=\linewidth]{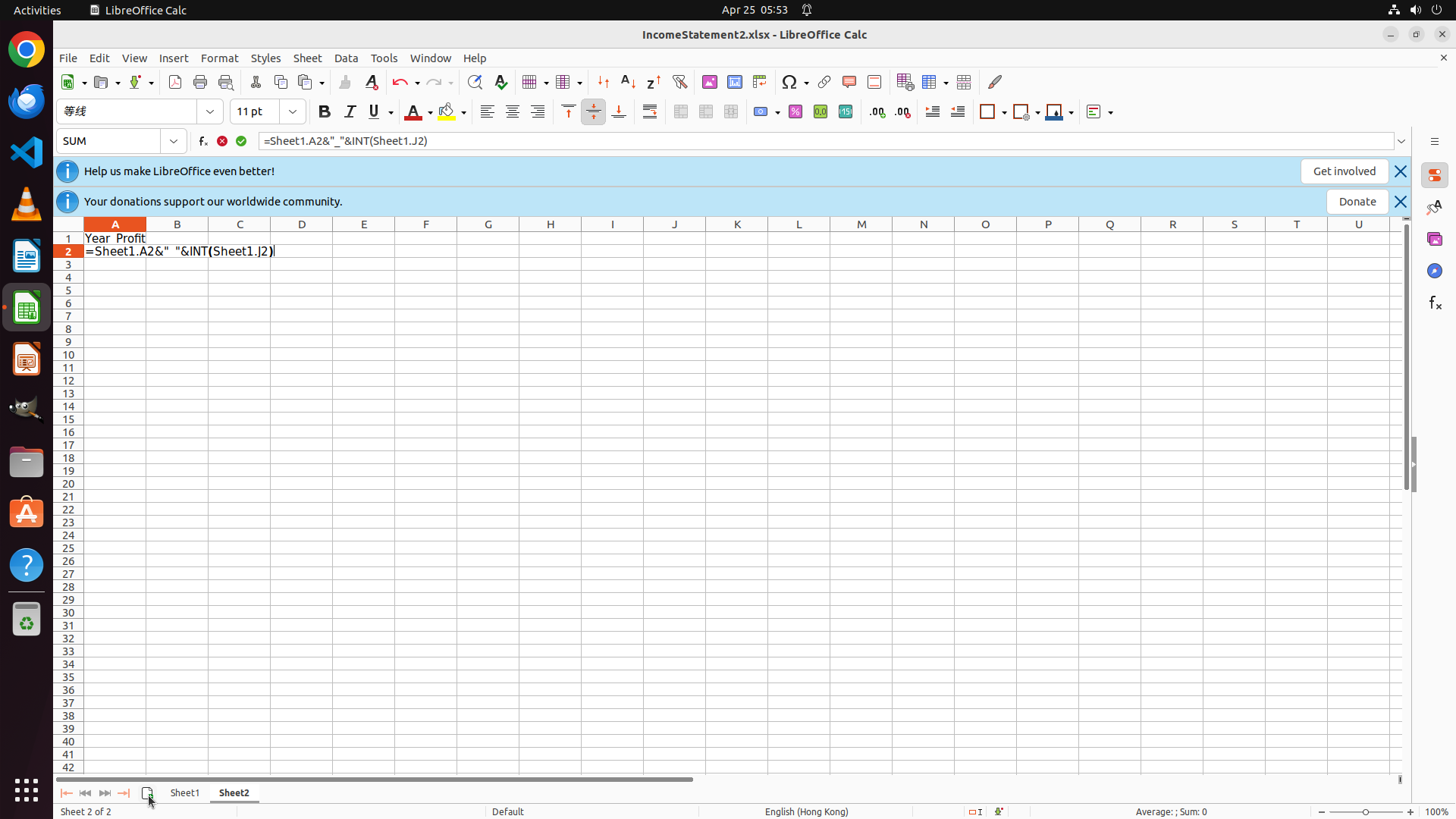} \\

    \scriptsize
    Entered the header \texttt{Year\_Profit} in \texttt{Sheet2!A1}.
    \par\texttt{Type(Year\_Profit)}
    &
    \scriptsize
    Confirmed the header and moved to the next row for formula entry.
    \par\texttt{Key(enter)}
    &
    \scriptsize
    Entered a direct row-wise formula, =Sheet1.A2\&"\_"\&INT(Sheet1.J2), and produced a concrete value
    in \texttt{A2}.
    \par\texttt{Type(=Sheet1.A2\&"\_"\&}
    \par\texttt{INT(Sheet1.J2))}
\end{tabularx}
\textit{The specialized model uses the simplest formula consistent with the task: a direct row-wise reference that immediately yields a valid \texttt{Year\_Profit} value.}
\end{tcolorbox}

\begin{tcolorbox}[
    colback=myyellow,
    colframe=black!55,
    boxrule=0.4pt,
    arc=2pt,
    left=4pt,
    right=4pt,
    top=3pt,
    bottom=3pt
]
\textbf{Analysis:}
The baseline model chooses an unnecessarily complex \texttt{VLOOKUP}-based
formula, repeatedly edits it, and eventually produces a \texttt{\#NAME?}
error. The specialized model instead uses a direct row-wise reference,
\texttt{=Sheet1.A2\&"\_"\&INT(Sheet1.J2)}, and then propagates it down the
column. The improvement is therefore not just a successful outcome but a
better solution strategy: it replaces brittle lookup logic with the simplest
reference pattern compatible with the task.
\end{tcolorbox}
\end{tcolorbox}
\caption{Case Study \#2 (Domain: \texttt{Libreoffice Calc})}
\end{figure}

\newpage

\begin{figure}[H]
\begin{tcolorbox}[
  colback=white,
  fontupper=\footnotesize,
  width=\textwidth,
  breakable
]
\textbf{Instruction}:
Move the title of page 2 to the bottom of the slide.

\vspace{0.5em}
\begin{tcolorbox}[colback=red!4, colframe=black!55, boxrule=0.4pt, arc=2pt, left=4pt, right=4pt, top=3pt, bottom=3pt, breakable]
\textbf{Before Specialization} \\
\vspace{0.3em}
\begin{tabularx}{\linewidth}{@{}YYY@{}}
    Step 43 & Step 45 & \textbf{Step 47} \\

    \includegraphics[width=\linewidth]{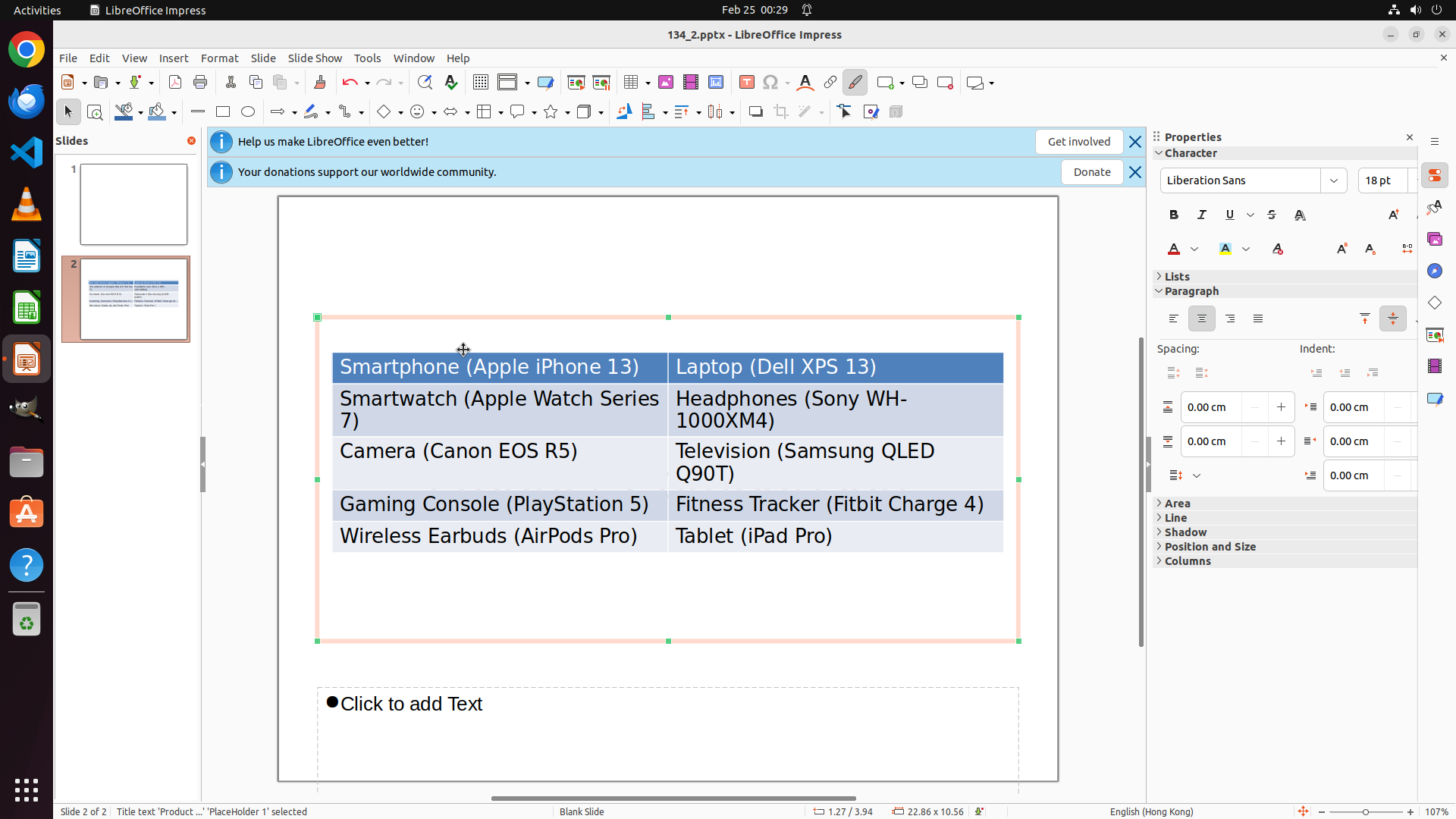}
    & \includegraphics[width=\linewidth]{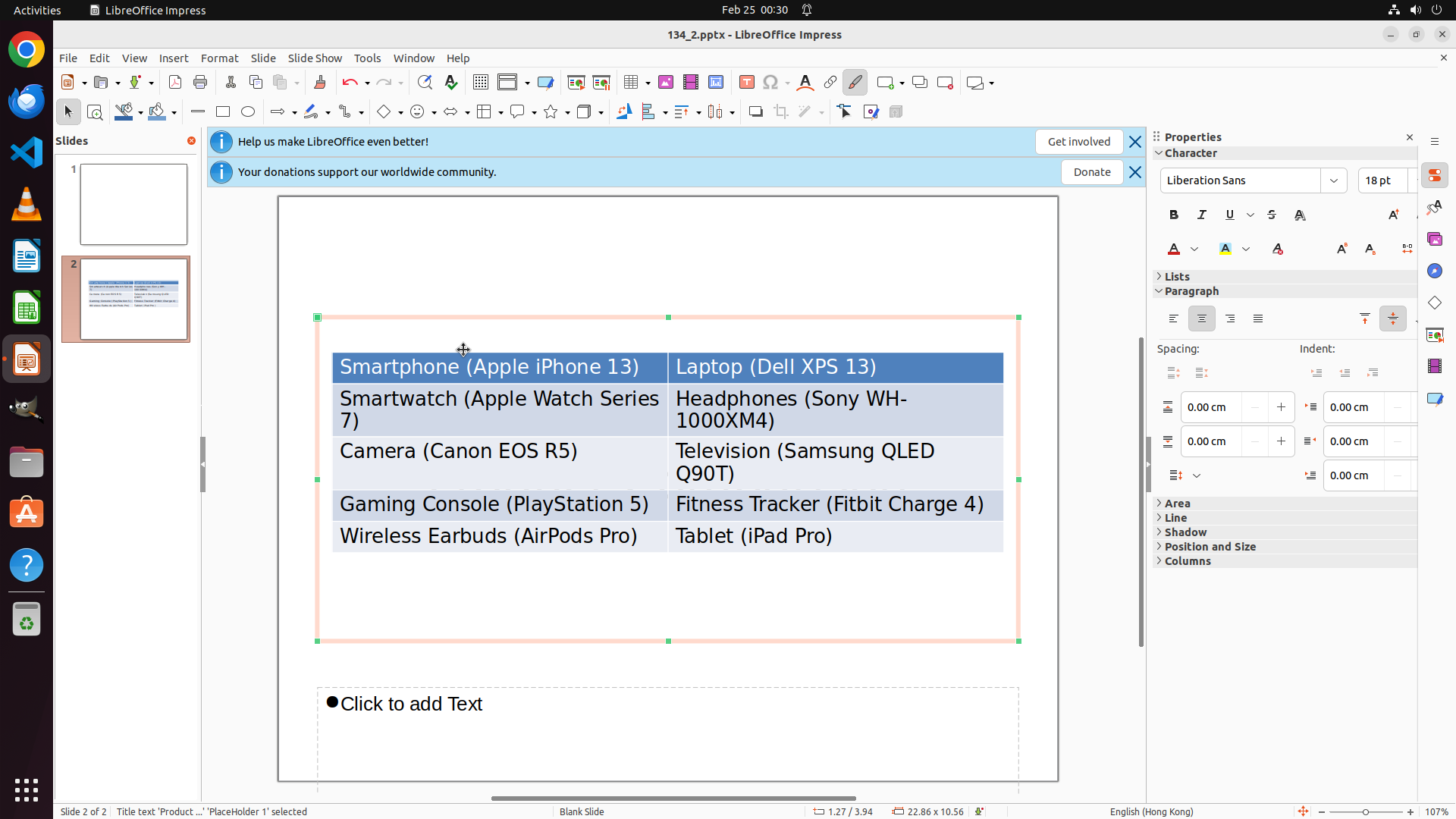}
    & \includegraphics[width=\linewidth]{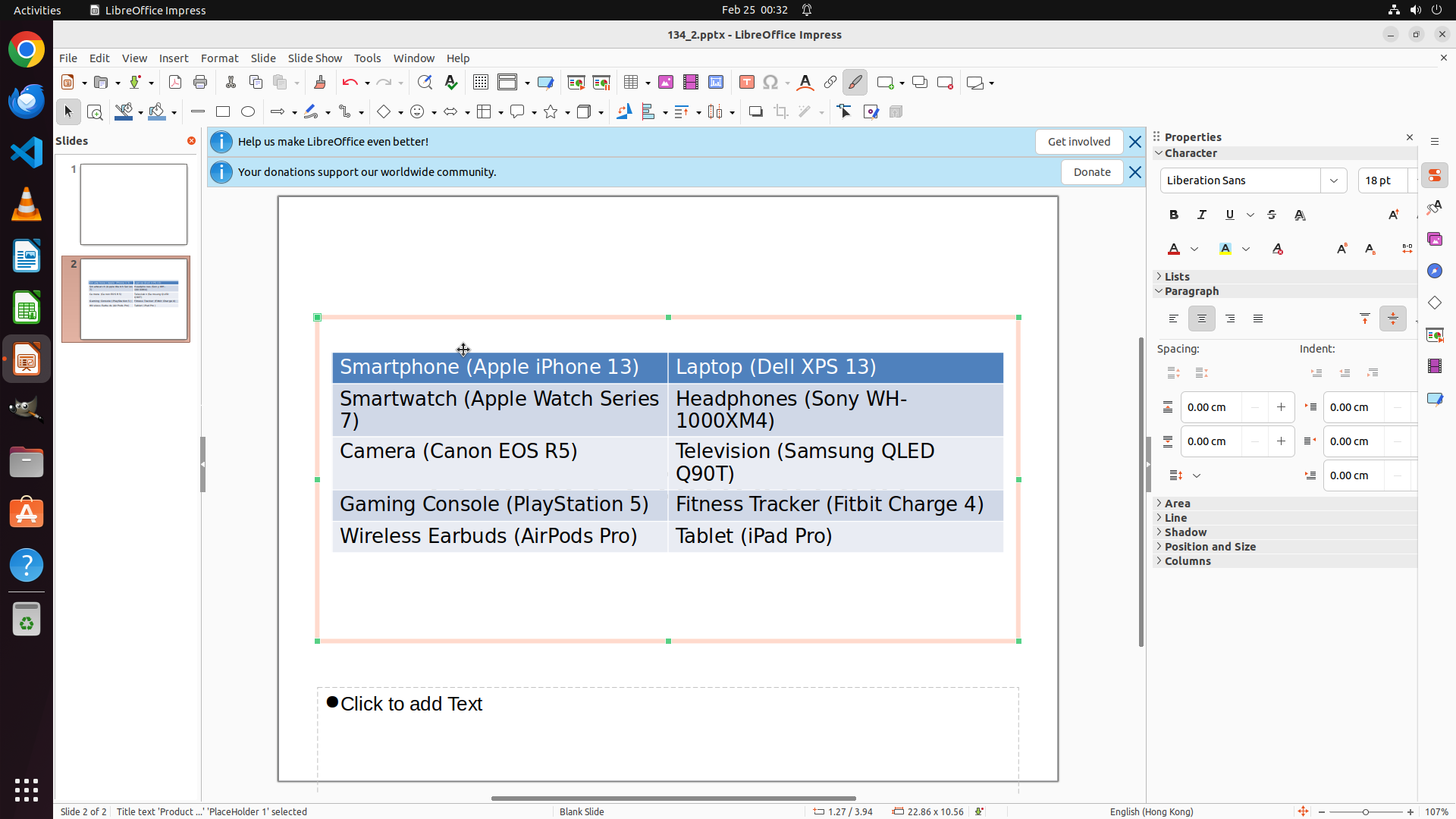} \\

    \scriptsize
    Repeatedly clicked on empty slide space in an attempt to clear the current
    selection before moving the title.
    \par\texttt{Click(599,449)}
    &
    \scriptsize
    Stayed stuck in the same deselection loop while the title remained near the
    top of the slide.
    \par\texttt{Click(599,449)}
    &
    \scriptsize
    Still attempted the same deselection action instead of re-entering a valid
    object-move workflow.
    \par\texttt{Click(599,449)}
\end{tabularx}

\textit{The baseline falls into a long recovery loop, repeatedly clicking empty slide space without returning to a successful object-move interaction.}
\end{tcolorbox}

\begin{tcolorbox}[colback=mygreen!8, colframe=black!55, boxrule=0.4pt, arc=2pt, left=4pt, right=4pt, top=3pt, bottom=3pt, breakable]
\textbf{After Specialization}

\vspace{0.3em}
\begin{tabularx}{\linewidth}{@{}YYY@{}}
    Step 16 & Step 18 & \textbf{Step 22} \\

    \includegraphics[width=\linewidth]{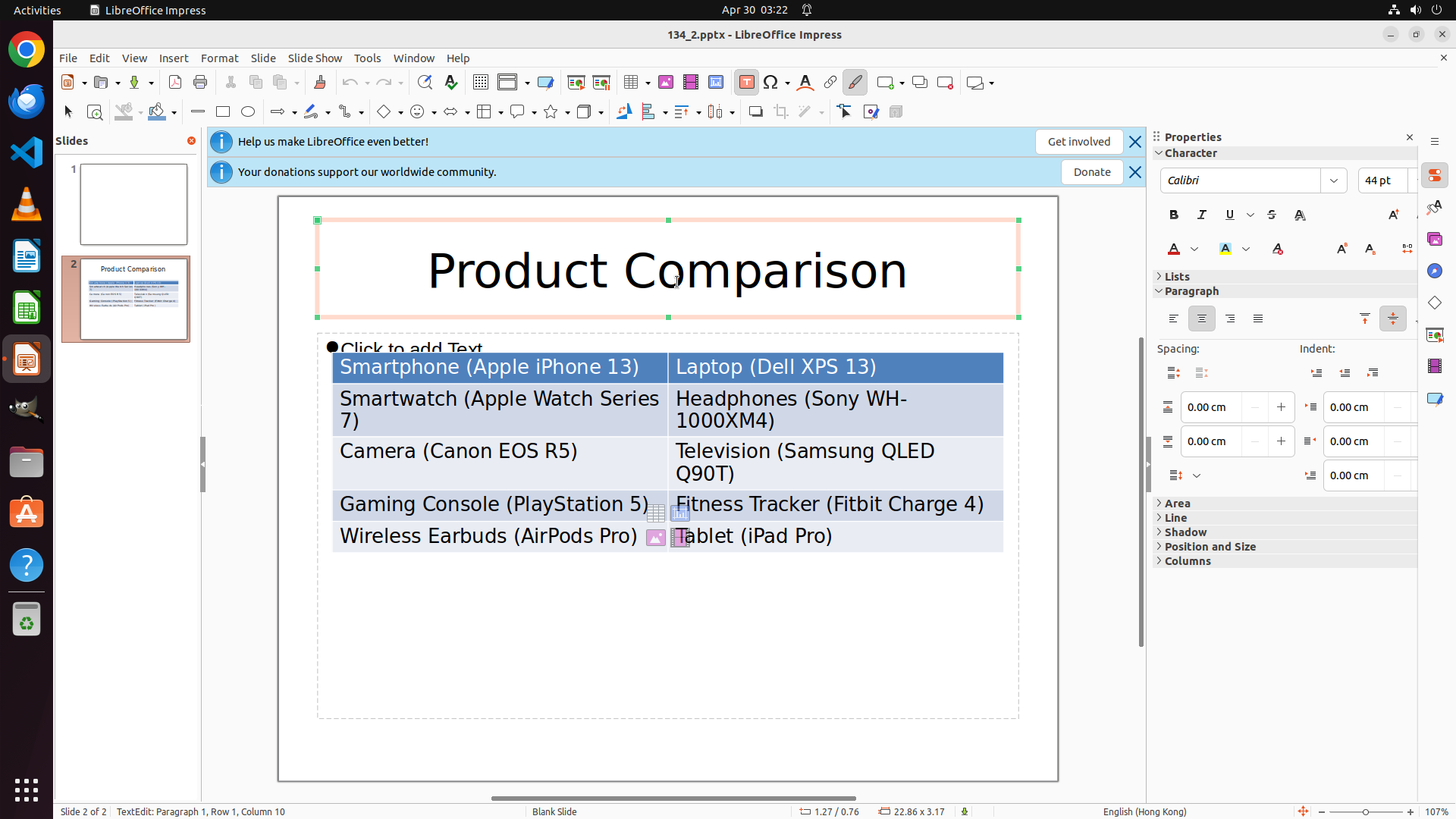}
    & \includegraphics[width=\linewidth]{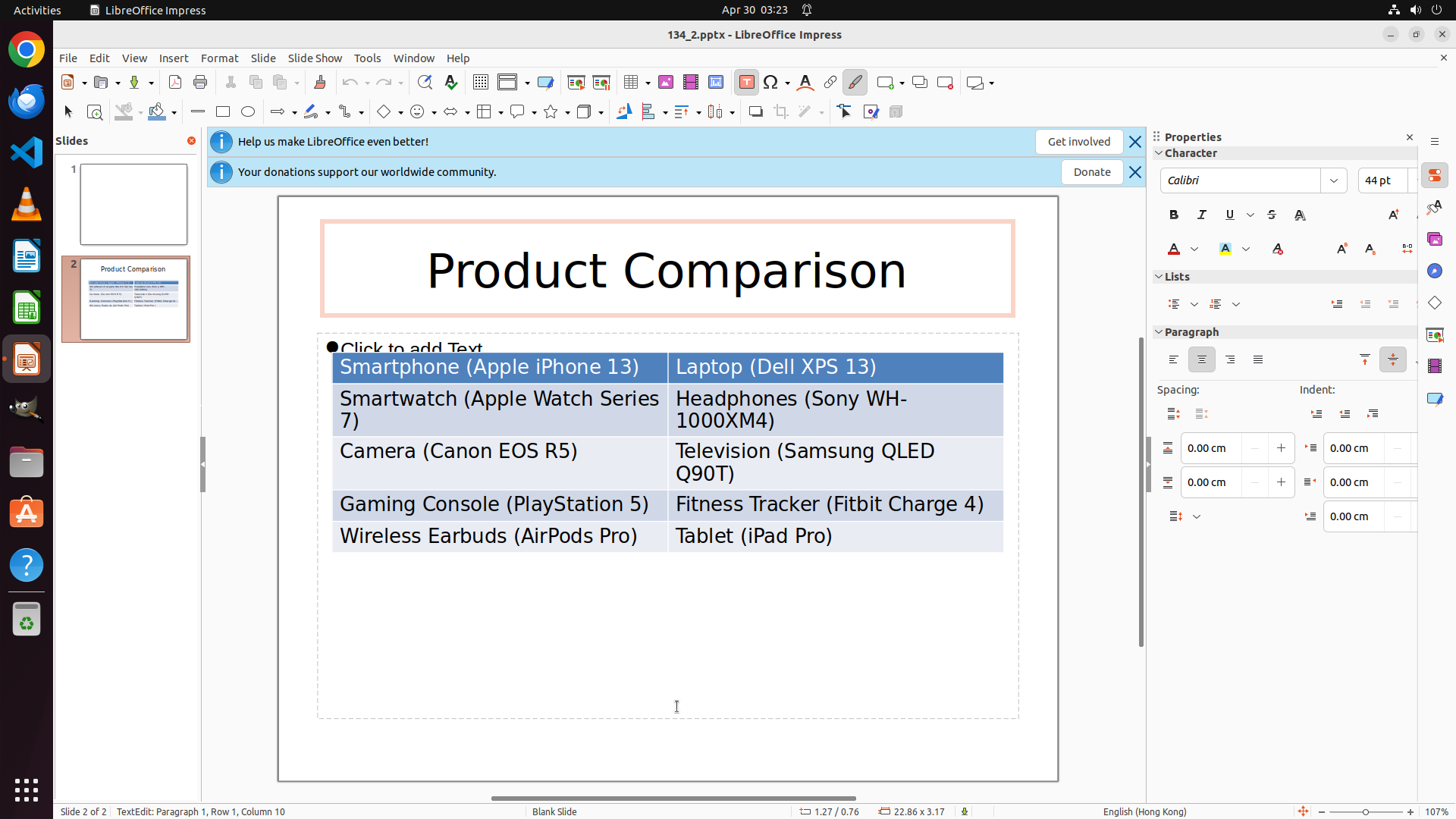}
    & \includegraphics[width=\linewidth]{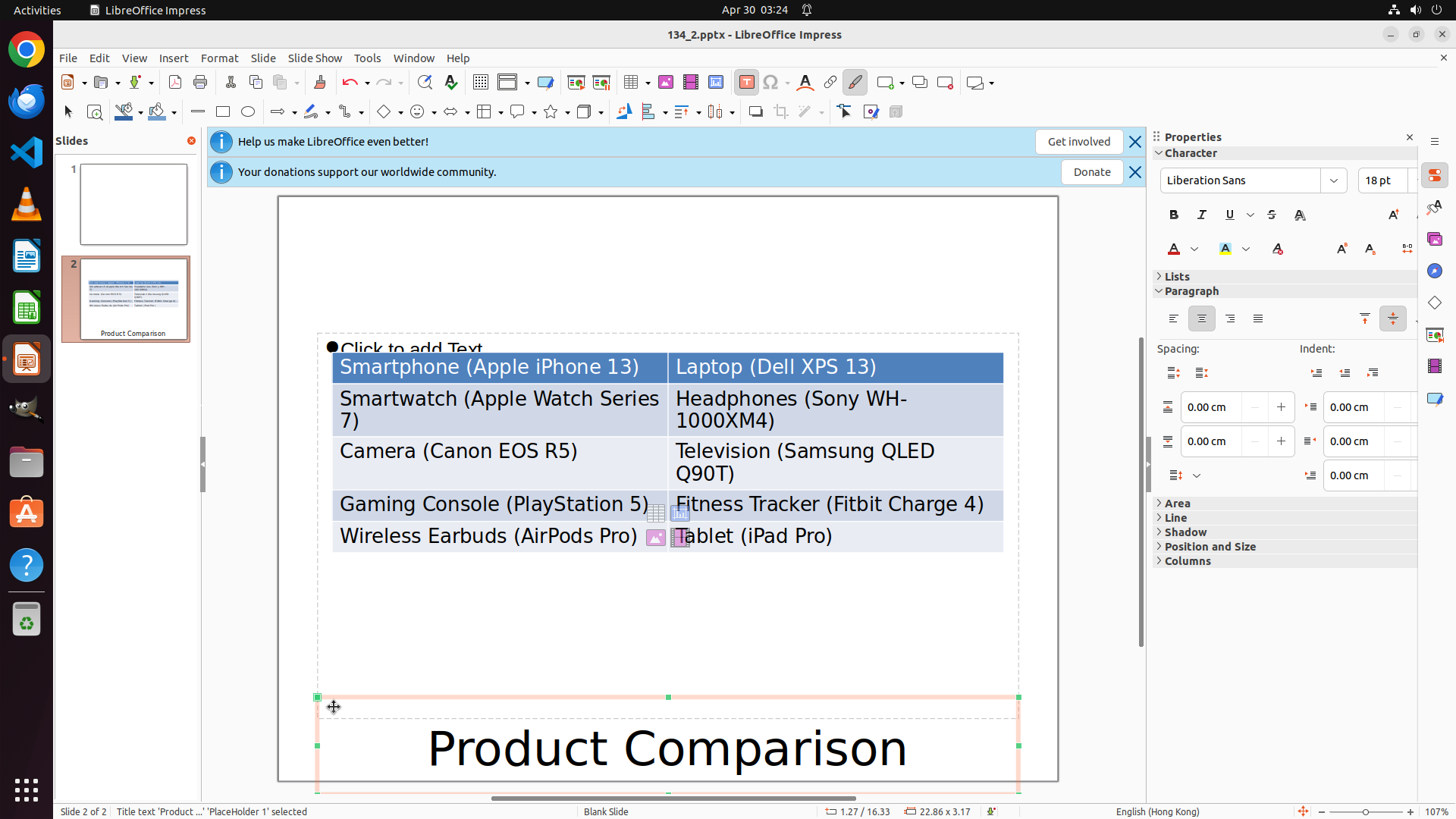} \\

    \scriptsize
    Explicitly exited text-edit mode after recognizing that direct dragging was
    blocked by the current interaction state.
    \par\texttt{Key(esc)}
    &
    \scriptsize
    Dragged the title box downward once object handles were visible again.
    \par\texttt{DragTo(882,920)}
    &
    \scriptsize
    Performed the final downward drag that places the title below the table at
    the bottom of the slide.
    \par\texttt{DragTo(428,920)}
\end{tabularx}
\textit{The specialized model recovers by leaving text-edit mode and then resumes the correct object-drag sequence until the title reaches the bottom region.}
\end{tcolorbox}

\begin{tcolorbox}[
    colback=myyellow,
    colframe=black!55,
    boxrule=0.4pt,
    arc=2pt,
    left=4pt,
    right=4pt,
    top=3pt,
    bottom=3pt
]
\textbf{Analysis:}
The core difference is recovery from an incorrect interaction mode. The
baseline remains trapped in a repeated deselection loop and never returns to a
successful move operation. The specialized model instead recognizes that it is
still in text-edit mode, exits that state with \texttt{Esc}, and then
completes the object-level drag needed to reposition the title.
\end{tcolorbox}
\end{tcolorbox}
\caption{Case Study \#3 (Domain: \texttt{Libreoffice Impress})}
\end{figure}

\newpage

\begin{figure}[H]
\begin{tcolorbox}[
  colback=white,
  fontupper=\footnotesize,
  width=\textwidth,
  breakable
]
\textbf{Instruction}:
Navigate to slide 5 and set the font color of all textboxes to yellow. Use
exactly `yellow'---no variations such as light yellow, dark yellow, or any
other color.

\vspace{0.5em}
\begin{tcolorbox}[colback=red!4, colframe=black!55, boxrule=0.4pt, arc=2pt, left=4pt, right=4pt, top=3pt, bottom=3pt, breakable]
\textbf{Before Specialization} \\
\vspace{0.3em}
\begin{tabularx}{\linewidth}{@{}YYY@{}}
    Step 7 & Step 9 & \textbf{Step 11} \\

    \includegraphics[width=\linewidth]{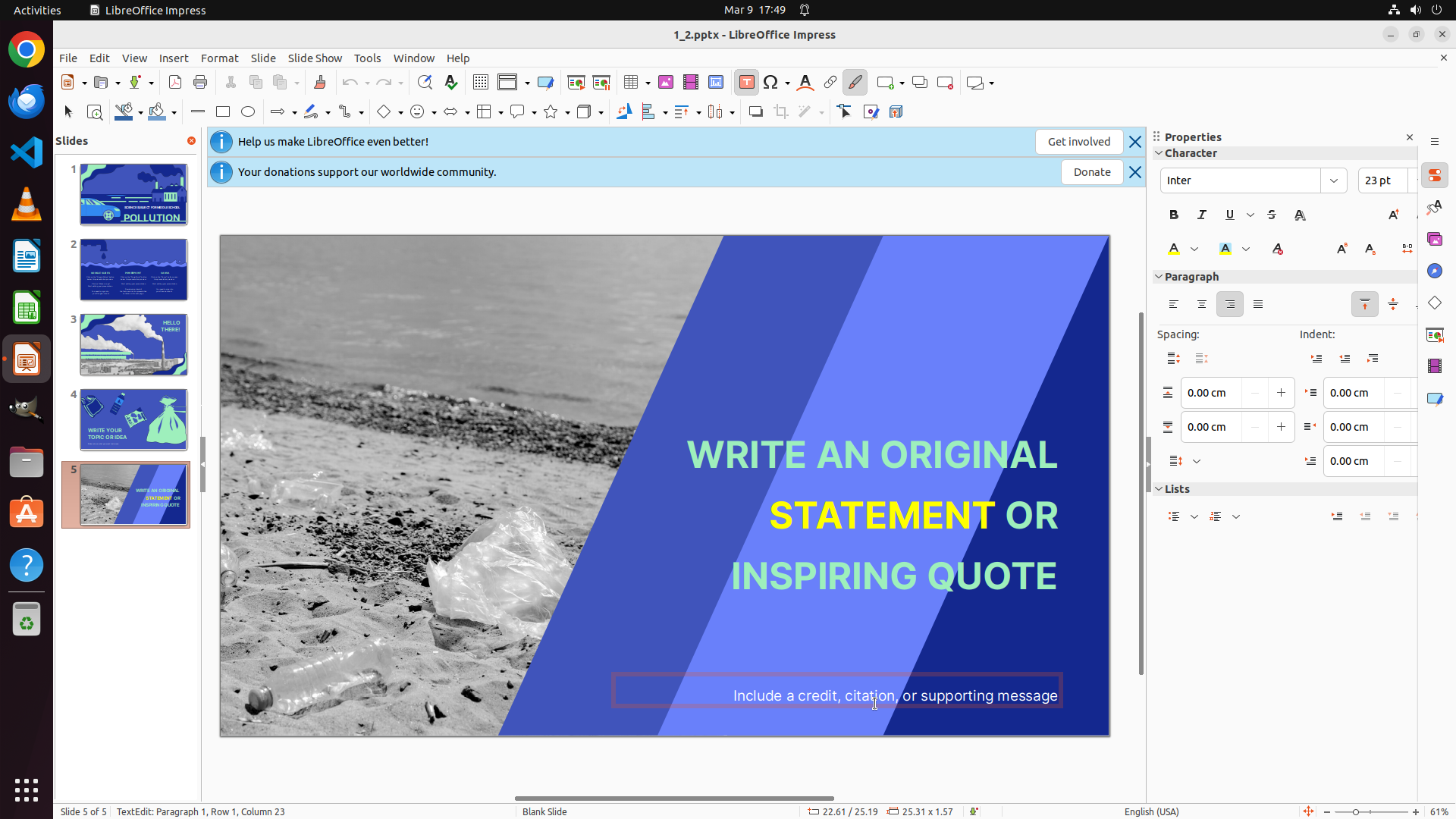}
    & \includegraphics[width=\linewidth]{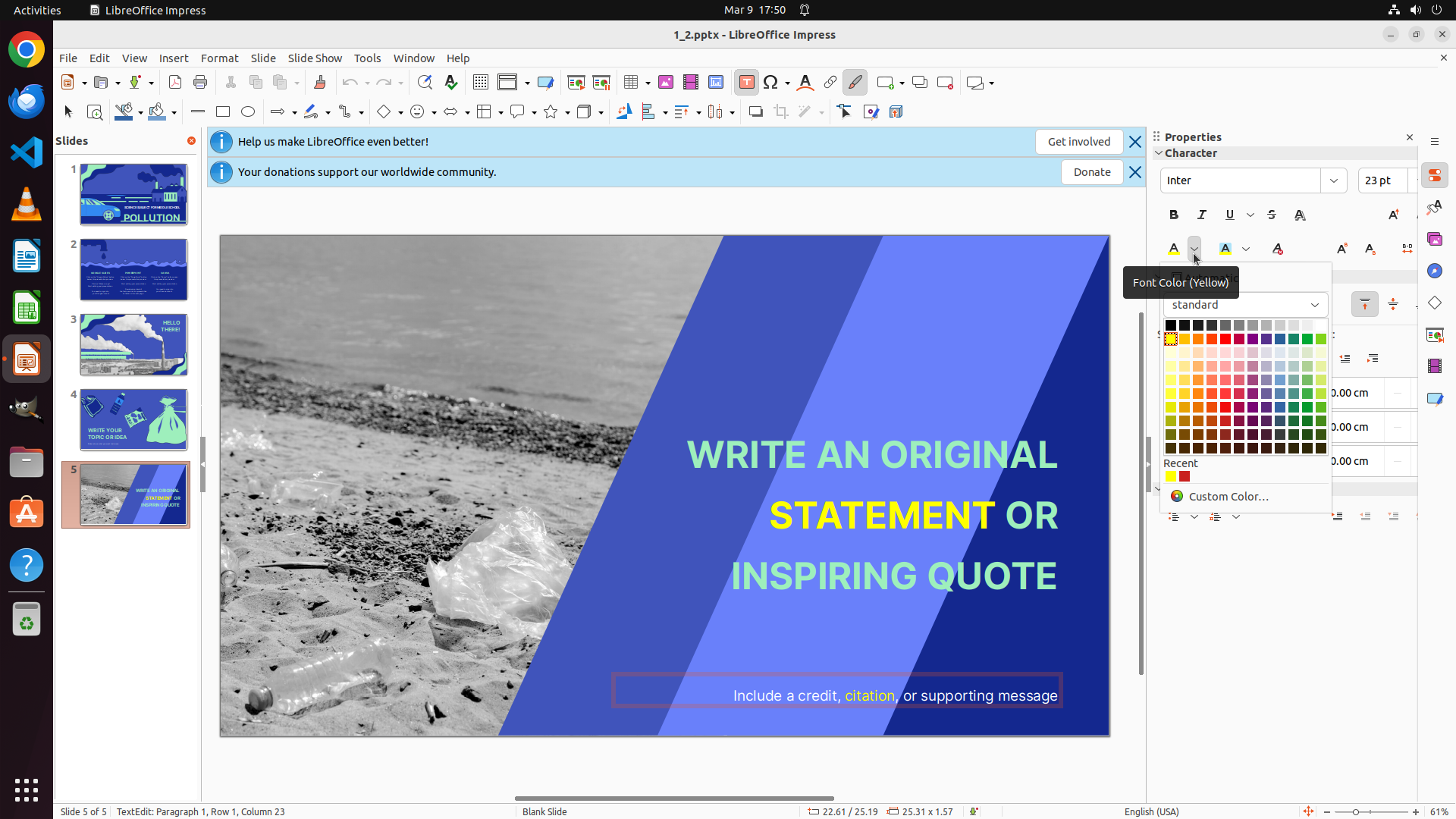}
    & \includegraphics[width=\linewidth]{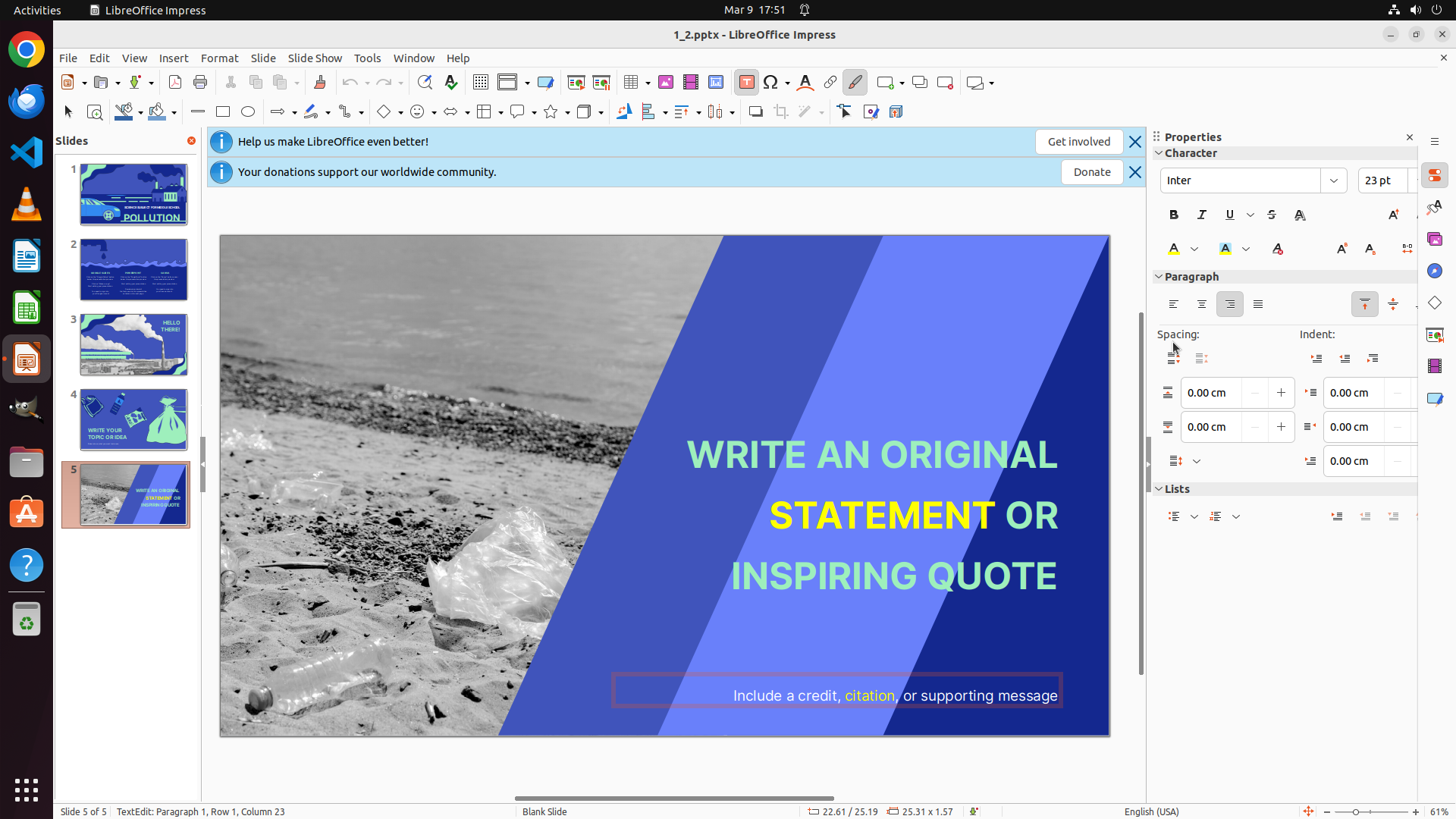} \\

    \scriptsize
    Switched from the title to the subtitle after recoloring only part of the
    slide text.
    \par\texttt{Click(1143,916)}
    &
    \scriptsize
    Opened the yellow color picker for the subtitle text.
    \par\texttt{Click(1570,330)}
    &
    \scriptsize
    Terminated once the title and subtitle were yellow, without covering the
    remaining text elements on the slide.
    \par\texttt{DONE}
\end{tabularx}

\textit{The baseline stops after recoloring two obvious textboxes, leaving the broader ``all textboxes'' requirement only partially satisfied.}
\end{tcolorbox}

\begin{tcolorbox}[colback=mygreen!8, colframe=black!55, boxrule=0.4pt, arc=2pt, left=4pt, right=4pt, top=3pt, bottom=3pt, breakable]
\textbf{After Specialization}

\vspace{0.3em}
\begin{tabularx}{\linewidth}{@{}YYY@{}}
    Step 26 & Step 28 & \textbf{Step 31} \\

    \includegraphics[width=\linewidth]{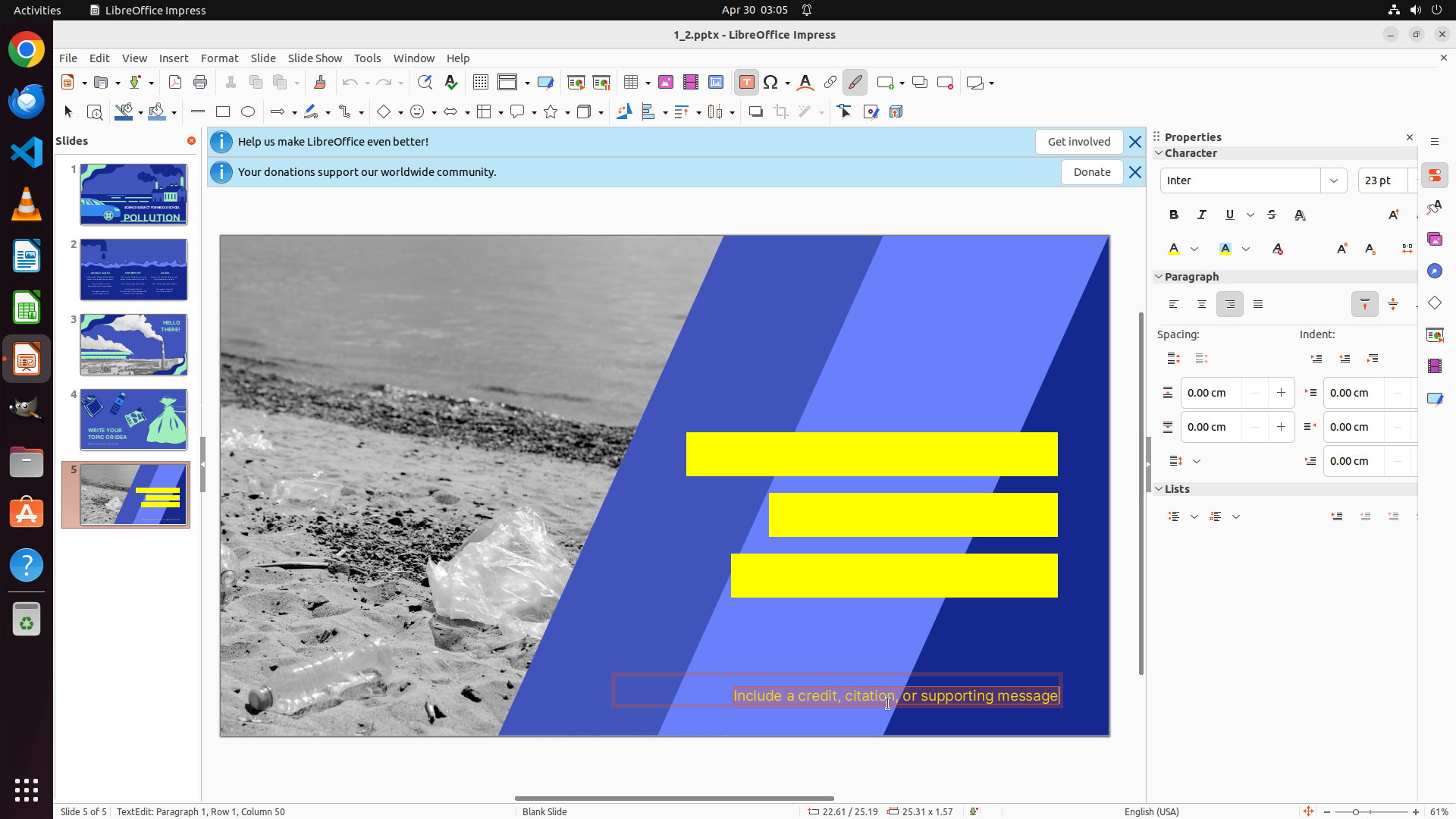}
    & \includegraphics[width=\linewidth]{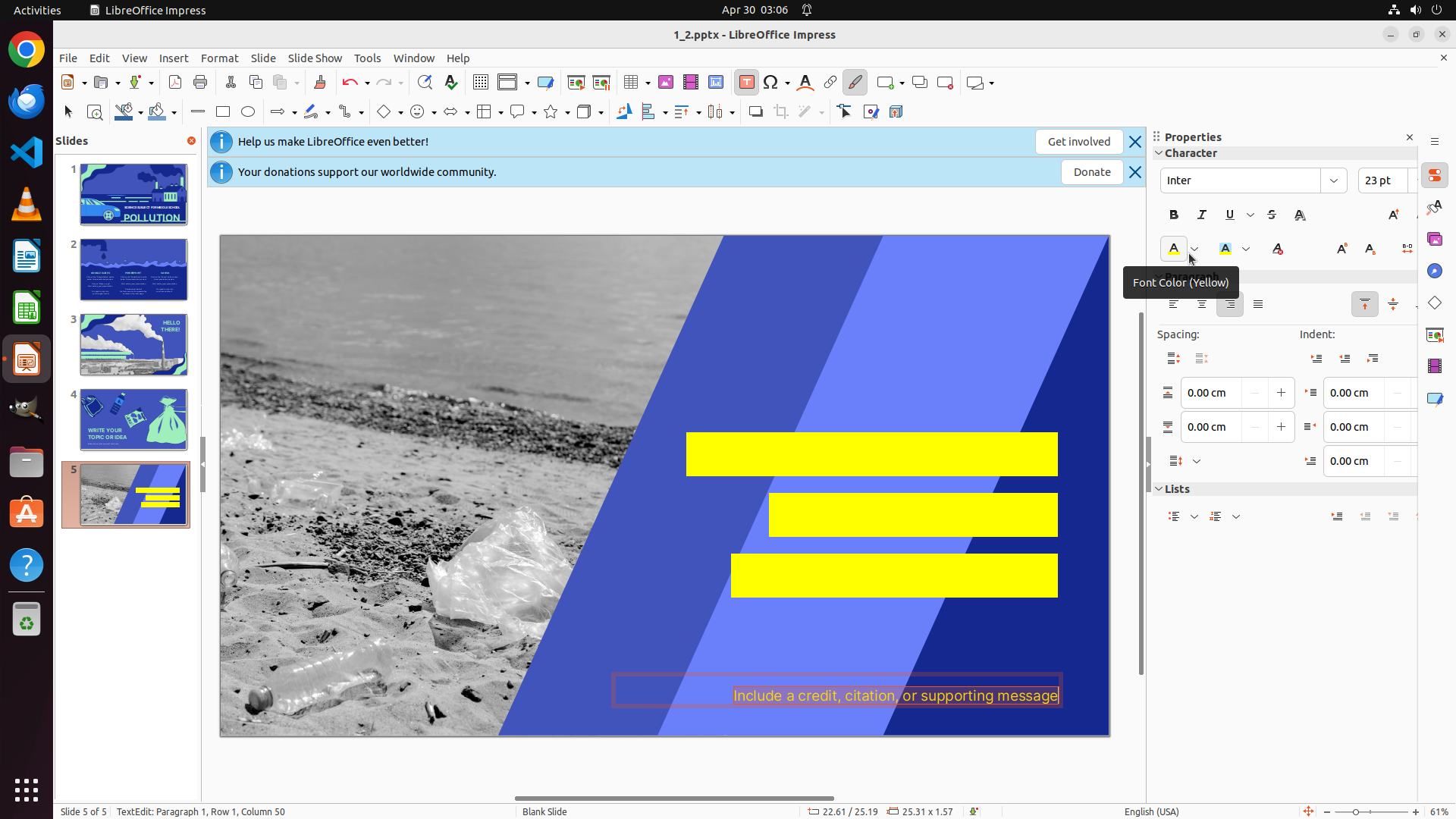}
    & \includegraphics[width=\linewidth]{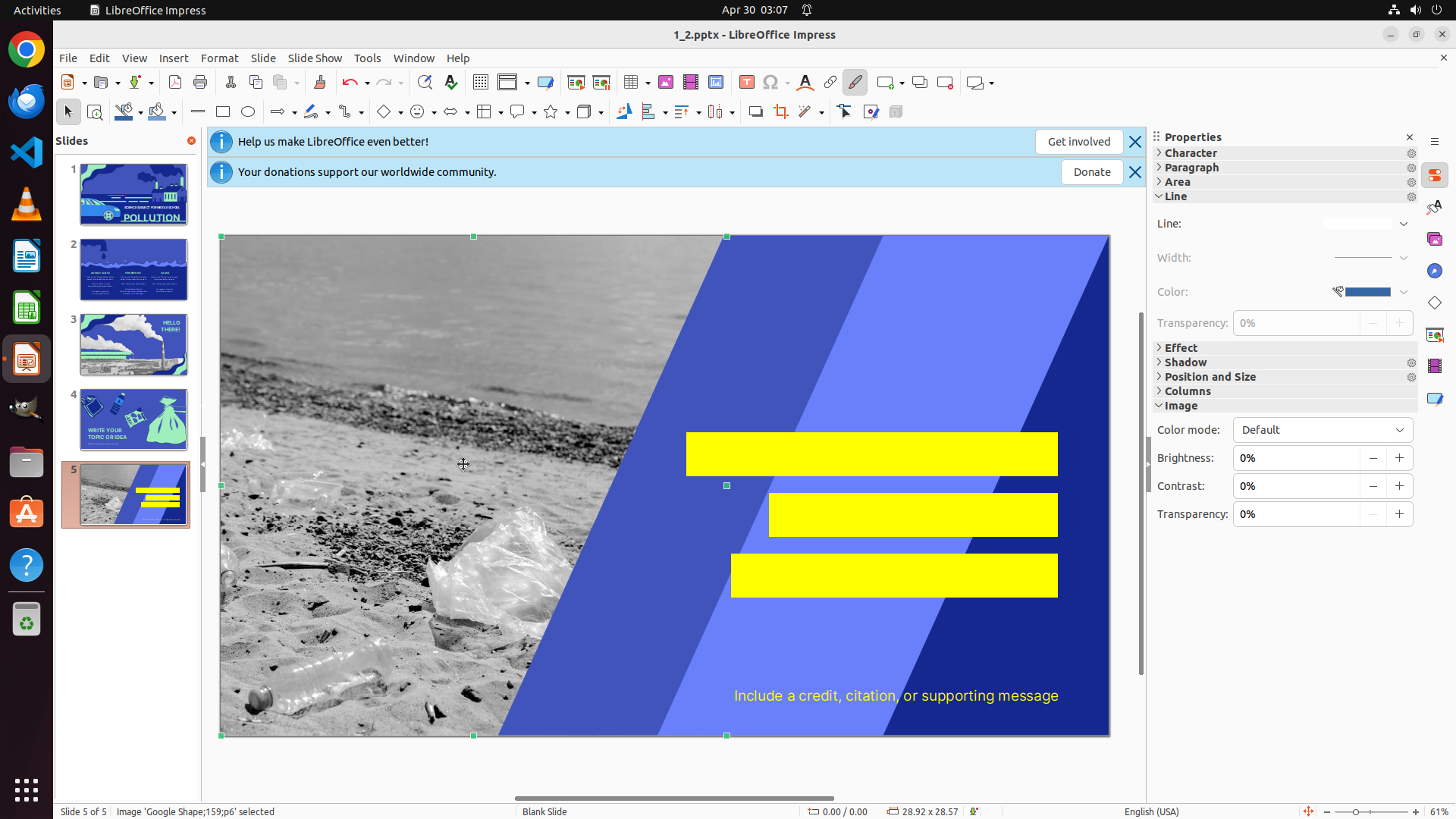} \\

    \scriptsize
    Re-selected the subtitle textbox and explicitly selected all text again to
    ensure the intended edit target was active.
    \par\texttt{HotKey(ctrl,a)}
    &
    \scriptsize
    Applied yellow through the font-color control after returning to the
    correct text-formatting context.
    \par\texttt{Click(1564,330)}
    &
    \scriptsize
    Continued until the full slide satisfied the instruction, including the
    remaining yellow bar text elements.
    \par\texttt{DONE}
\end{tabularx}
\textit{The specialized model continues beyond the first successful edits and only finishes once all required text elements on slide 5 are yellow.}
\end{tcolorbox}

\begin{tcolorbox}[
    colback=myyellow,
    colframe=black!55,
    boxrule=0.4pt,
    arc=2pt,
    left=4pt,
    right=4pt,
    top=3pt,
    bottom=3pt
]
\textbf{Analysis:}
This example highlights improved task-completion coverage rather than a single
local operation. The baseline performs the obvious recoloring steps and then
terminates early. The specialized model instead keeps checking whether the
entire slide matches the instruction, re-enters the text-formatting context
when needed, and finishes only after all required textboxes are yellow.
\end{tcolorbox}
\end{tcolorbox}
\caption{Case Study \#4 (Domain: \texttt{Libreoffice Impress})}
\end{figure}



\end{document}